\documentclass[]{gtech}
\usepackage{amssymb}
\usepackage{graphicx}
\usepackage{multirow}
\usepackage{bigdelim}
\usepackage{todonotes}
\usepackage{longtable}
\usepackage{tabularray}
\usepackage{wrapfig}
\usepackage[most]{tcolorbox} 
\usepackage{xcolor}
\usepackage{url}
\usepackage{float}
\usepackage{caption}
\usepackage{CJKutf8}
\usepackage[most]{tcolorbox}  
\usepackage{xcolor}
\usepackage{enumitem}          
\usepackage{academicons}
\usepackage{fontawesome5}

\renewcommand{\title}[1]{\newcommand{\titlelist}{{\huge\fontfamily{optimistic}\selectfont #1}}}

\newcommand{\modelraw}{Ming-UniAudio}

\newcommand{\model}{\texttt{Ming-UniAudio}}

\newcommand{\modeledit}{\texttt{Ming-UniAudio-Edit}}
\newcommand{\editbenchmark}{\texttt{Ming-Freeform-Audio-Edit}}
\newcommand{\mingaudiotokenizer}{\texttt{MingTok-Audio}}

\definecolor{prompt}{HTML}{5f84e4}
\definecolor{img}{HTML}{820100}

\usepackage{arydshln}

\definecolor{CQColor}{rgb}{0.0,0.0,1.0} 

\definecolor{TSColor}{rgb}{0.5,0.0,0.8} 

\definecolor{CQRColor}{rgb}{1.0,0.0,1.0} 

\usepackage{colortbl}
\usepackage{amssymb}
\usepackage{pifont}
\usepackage{booktabs,multirow}
\usepackage{makecell}
\usepackage{tabulary}
\usepackage{fontawesome5}
\usepackage{bbding}
\usepackage{multicol}

\newlength\savewidth

\title{\textcolor{blue}{Ming-UniAudio}: \\Speech LLM for Joint Understanding, Generation and Editing with Uni{f}ied Representation }

\author[*]{Inclusion AI, Ant Group}
\contribution[*]{See Contributions section (Sec.~\ref{sec:contri}) for full author list.}

\abstract{\fontsize{11pt}{12pt} \textit{
Existing speech models suffer from competing requirements on token representations by understanding and generation tasks. This discrepancy in representation prevents speech language models from performing instruction-based free-form editing. To solve this challenge, we introduce a novel framework that unifies speech understanding, generation, and editing. The core of our unified model is a unified continuous speech tokenizer \textbf{\mingaudiotokenizer}, the first continuous tokenizer to effectively integrate semantic and acoustic features, which makes it suitable for both understanding and generation tasks. Based on this unified continuous audio tokenizer, we developed the speech language model \textbf{\model{}}, which achieved a balance between generation and understanding capabilities. \textbf{\model{}} sets new state-of-the-art (SOTA) records on 8 out of 12 metrics on the ContextASR benchmark. Notably, for Chinese voice cloning, it achieves a highly competitive Seed-TTS-WER of 0.95. Leveraging this foundational model, we further trained a dedicated speech editing model \textbf{\modeledit{}}, the first speech language model that enables universal, free-form speech editing guided solely by natural language instructions, handling both semantic and acoustic modifications without timestamp condition. To rigorously assess the editing capability and establish a foundation for future research, we introduce \textbf{\editbenchmark{}}, the first comprehensive benchmark tailored for instruction-based free-form speech editing, featuring diverse scenarios and evaluation dimensions spanning semantic correctness, acoustic quality, and instruction alignment. We open-sourced the continuous audio tokenizer, the unified foundational model, and the free-form instruction-based editing model to facilitate the development of unified audio understanding, generation, and manipulation.
}}

\date{Oct 16, 2025\vspace{-1mm}}
\gtechdata[Project Homepage]{\href{https://xqacmer.github.io/Ming-Unitok-Audio.github.io}{\underline{Link}}\vspace{-1mm}}
\gtechdata[Code]{\url{https://github.com/inclusionAI/Ming-UniAudio}}
\gtechdata[Model]{\url{https://huggingface.co/inclusionAI}}

\begin{document}
\maketitle

\section{Introduction}
\label{sec:intro}

The rapid breakthroughs in large language models (LLMs) \citep{Brown2020LanguageMA,Yang2025Qwen3TR} have significantly improved the capabilities of speech foundation models in both understanding \citep{chu2023qwenaudio,Wu2025StepAudio2T} and generation \citep{Wang2023NeuralCL,CosyVoice,jia2025ditar} . These advances enable more accurate and fine-grained comprehension of spoken language, as well as more fluent and natural speech output for human-machine interaction \citep{Xu2025Qwen3OmniTR}. 
In terms of speech representations, understanding and generation tasks impose fundamentally different demands: understanding benefits from compact, semantics-focused encodings, while generation requires rich, high-quality acoustic details.As a result, most existing speech language models resort to one of two architectural compromises: either maintaining separate representations for understanding and generation \citep{Xu2025Qwen3OmniTR,KimiTeam2025KimiAudioTR}, or relying on discrete tokens for both \citep{defossez2024moshi, coreteam2025mimoaudio,Huang2025StepAudioUU}. 
The former approach, separated representation, presents a fundamental challenge for editing, as the disjointed representations for understanding and generation prevent the seamless workflow required to interpret an instruction and synthesize the modified audio \citep{Le2023VoiceboxTM}. Whereas the latter approach, discrete representation, suffers from loss of speech details due to quantization. In terms of speech tasks, there is currently no single model that is capable of performing semantic and acoustic editing on input speech with free-form instructions. In contrast, multi-round editing has already been well demonstrated in the image domain, such as GPT-4o, Gemini-Image-Flash 2.5 \citep{openai2024gpt4o,google2025gemini}. To reconcile the issue of competing requirements of representations on understanding and generation tasks, we propose a VAE-based continuous audio tokenizer that unifies semantic and acoustic representation. Furthermore, we introduce a novel capability, free-form speech editing, which enables fine-grained and high-quality speech manipulation. However, building such a unified system requires addressing several fundamental challenges, which we detail in the following.


\begin{itemize}
\item \textbf{Inconsistent Representations: Understanding vs. Generation:} 
A primary hurdle lies in inconsistent representational needs for speech understanding and generation. Understanding tasks, such as automatic speech recognition (ASR) and sentiment analysis, primarily require high-level semantic information. In contrast, speech generation (e.g., text-to-speech, voice conversion) demands faithful reconstruction of fine-grained acoustic details, including prosody, timbre, and intonation. It is very challenging to reconcile their distinct requirements within a single representation, often leading to models that excel in one aspect at the expense of the other.

\item \textbf{Different Training Objectives}:  
In the joint training of understanding and generation, there are several challenges. The optimization objectives for the two tasks are distinct, and they present different levels of training difficulty. Moreover, generation tasks typically require more training iterations. As a result, it remains a pressing problem to determine the training ratio between understanding and generation tasks to simultaneously achieve optimal performance for both tasks, especially under the multi-stage training design.

\item \textbf{Limitations of Speech Editing:}
Existing approaches to speech editing are generally restrictive. Many require complex, multi-stage pipelines or operate on a limited set of predefined operations (e.g., denoising, speed alteration). Expanding into more complex `free-form' editing presents further challenges. Semantic content modifications (e.g., insertion, deletion, replacement) often demand timestamp conditions to specify the edit location. The lack of truly instruction-guided free-form editing capabilities limits the practical utility and accessibility of current speech editing tools.

\end{itemize}

\label{sec:method}
\begin{figure}[t]
    \centering
    \includegraphics[width=0.8\linewidth]{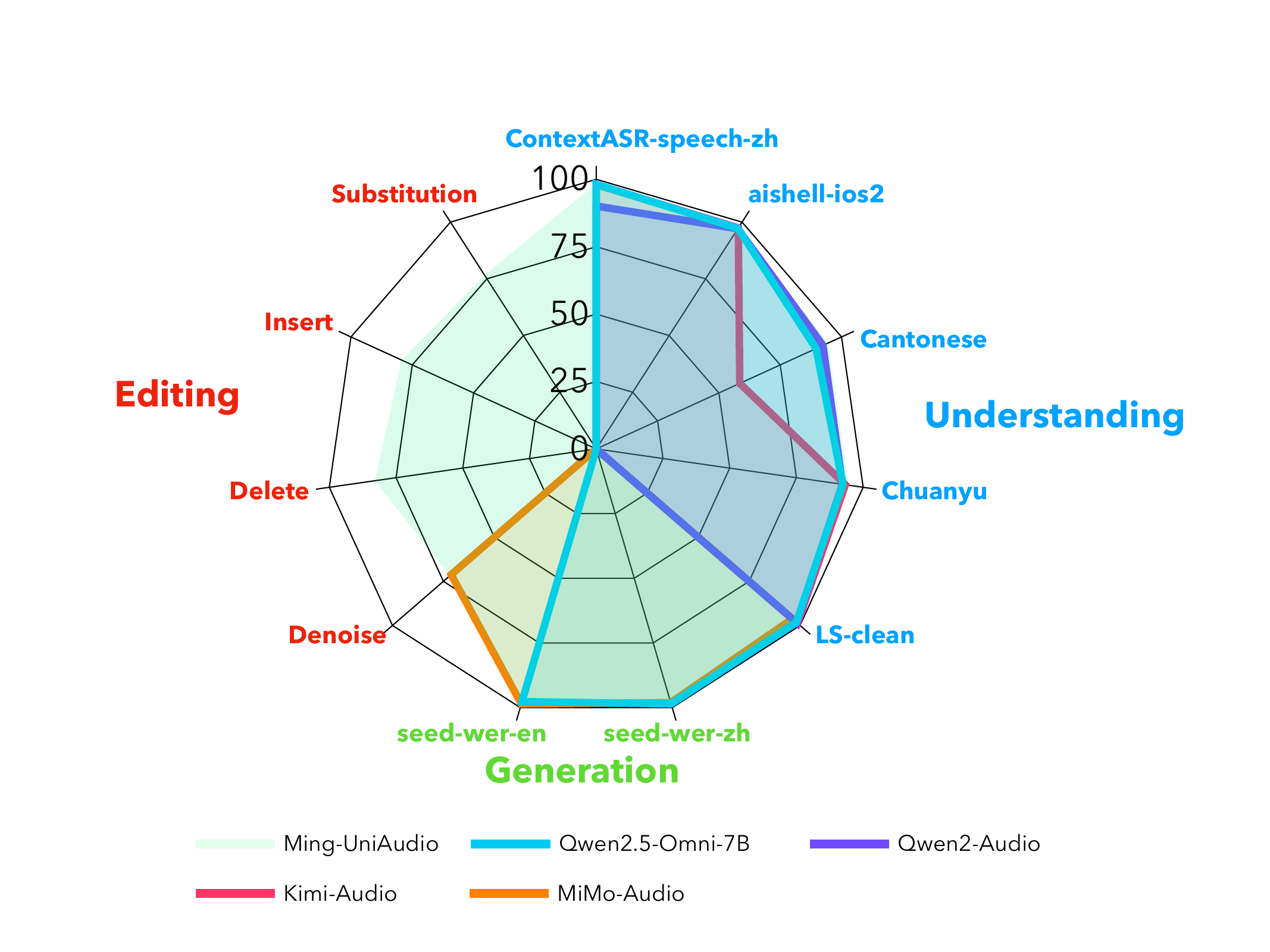}
    \caption{Comparison of \model{} with Open-Source Speech LLMs across Understanding, Generation, and Editing Tasks. }
    \label{radar_score}
    \vspace{-4mm}
\end{figure}


To address these issues, we propose \model{}, a novel framework designed to achieve unified speech understanding, generation, and free-form editing. Our key contributions are as follows.

\begin{itemize}
\item \textbf{Unified Continuous Speech Tokenizer:} We introduce a novel unified continuous speech tokenizer, which is built on a Variational Autoencoder (VAE) framework and implemented with a purely causal transformer architecture. This tokenizer unifies semantic and acoustic representations, optimized by multistage training and semantic distillation from an LLM. It simultaneously extracts high-dimensional unified features $Z_\text{uni}$ as a universal interface for LLMs and low-dimensional acoustic latent vector $Z_\text{latent}$ crucial for high-quality generation. Furthermore, this latent vector can be transformed into high-dimensional unified features via a semantic module, thereby forming a complete closed loop with the LLM's understanding and generation tasks.

\item \textbf{Unified Speech Language Model for Generation and Understanding:} We pretrained an end-to-end unified speech language model capable of both understanding and generation with a single LLM backbone. This model incorporates a per-token diffusion head that takes the LLM's hidden states and predicted semantic sequences as conditions, ensuring high-quality and natural speech synthesis.

\item \textbf{Instruction-Guided Free-Form Speech Editing:} 
To the best of our knowledge, this is the first work to realize a truly universal speech editing framework based entirely on free-form natural language instructions. This framework enables comprehensive editing, covering both semantic content modifications (insertion, deletion, substitution) and acoustic attribute adjustments (denoising, speed/pitch/volume, dialect/emotion conversion, etc.) without requiring the user to specify an explicit edit region. We propose Audio-Edit-Benchmark, the first open-source free-form evaluation set comprising editing tasks of three semantic and five acoustic types, to evaluate the model's editing performance.

\end{itemize}

\section{Related Work}

\subsection{Speech Tokenizer}
Speech tokenization is crucial for representing continuous speech signals. Acoustic tokens, such as SoundStream \citep{zeghidour2021soundstream} or EnCodec \citep{defossez2022highfi}, excel at preserving high reconstruction quality but often lack rich semantic information, which can result in compromised pronunciation stability \citep{Wang2023NeuralCL}. Conversely, semantic tokens, such as HuBERT, Whisper, and CosyVoice \citep{Hsu2021HuBERTSS, Whisper, du2024cosyvoice}, are rich in semantic content but tend to lose acoustic details. As a consequence, a separate diffusion model is required to convert these tokens into a Mel spectrogram and inject acoustic features like timbre, prosody, and style. To address the aforementioned issues, SpeechTokenizer \citep{zhang2023speechtokenizer} uses knowledge distillation to integrate semantic and acoustic features, and Step Audio Tokenizer \citep{Huang2025StepAudioUU} attempts to address this by interleaving semantic and acoustic tokens. However, their discrete nature makes them suboptimal for understanding-based tasks such as speech recognition. Inspired by the work of MingTok-Vision \citep{Huang2025MingUniVisionJI} in comupter vision, we present a VAE-based continuous tokenizer with unified semantic and acoustic representation, offering greater performance than discrete approaches. This unified approach shows that a future with seamless, free-form speech editing is now feasible. This capability is significant because it bridges the critical gap between understanding and generation.

\subsection{Unified Speech Understanding and Generation Models}
Recent efforts have aimed to build unified models for speech understanding and generation. KimiAudio \citep{KimiTeam2025KimiAudioTR} proposes an audio tokenizer that concatenates the semantic audio token with continuous acoustic vectors from Whisper \citep{Whisper} encoder to enhance perception capability and output a discrete semantic token. DualSpeechLM \citep{Wang2025DualSpeechLMTU} proposes a dual token modeling paradigm, using a dedicated USToken for input and an acoustic token for output. Other models like SpeechGPT \citep{zhang2024speechgpt}, and Moshi \citep{defossez2024moshi} have explored different strategies, often relying on large-scale paired data or exhibiting limitations in truly unified representation. Our unified tokenizer offers a more inherent and continuous representation of semantic and acoustic information, which is effective for both generation and understanding tasks. Furthermore, our token-level autoregressive diffusion head for generation is hypothesized to yield higher generation quality compared to a discrete token followed by a sentence-level diffusion module, which has been proved by DitAR \citep{jia2025ditar} and VibeVoice \citep{Peng2025VibeVoiceTR}.

\subsection{Instruction-Guided Speech Editing}
The field of speech editing has seen growth, particularly with instruction-guided approaches. VoiceBox \citep{Le2023VoiceboxTM} introduced a non-autoregressive editing paradigm based on flow matching. However, its reliance on alignment models like MFA to specify the editing region adds operational overhead in practical scenarios. InstructSpeech \citep{huang2024instructspeech} represents a significant advancement, employing triplet data construction, multitask learning, and multistep reasoning to enable instruction-based editing. However, when free-form instructions are given, a timestamp alignment mechanism is still required to locate the specified region. Other models, like VoiceCraft \citep{Peng2024VoiceCraftZS} and EdiTTS \citep{Tae2021EdiTTSSE}, often have limitations in truly "free-form" capabilities, either requiring explicit region masking or being restricted to specific types of edits.  Furthermore, paralinguistic tasks, including emotion and accent conversion, require that a speech LLM possesses strong comprehension abilities. Our framework directly addresses this by enabling comprehensive, unconstrained editing based solely on natural language instructions.

\section{Model Architecture }

\subsection{Unified Continuous Tokenizer}
Our unified streaming tokenizer is the cornerstone of \model, designed to resolve the representation inconsistency between understanding and generation by simultaneously extracting semantic and acoustic features in a streamable manner.
\subsubsection{Design Principles and Rationale}
\label{sec:tokenizer_design}

\begin{figure}[htb]
    \centering
    \begin{subfigure}{\textwidth}
        \includegraphics[width=0.9\linewidth]{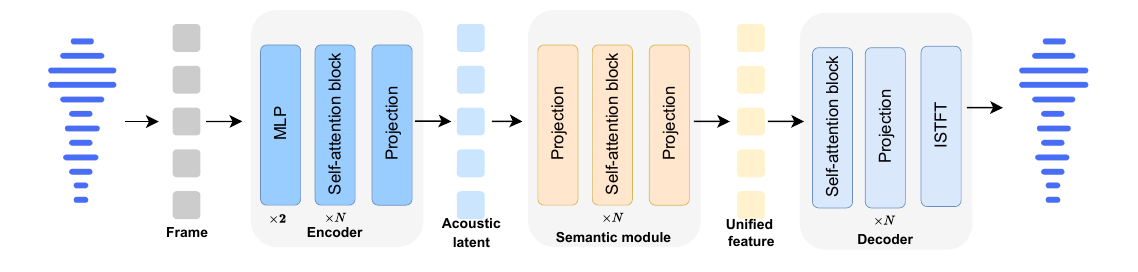}
        \vspace{-2mm}
        \caption{The architecture of \mingaudiotokenizer.}
        \label{Ming_unitok}
        \vspace{2mm}
    \end{subfigure}

    \begin{subfigure}{\textwidth}
        \includegraphics[width=0.9\linewidth]{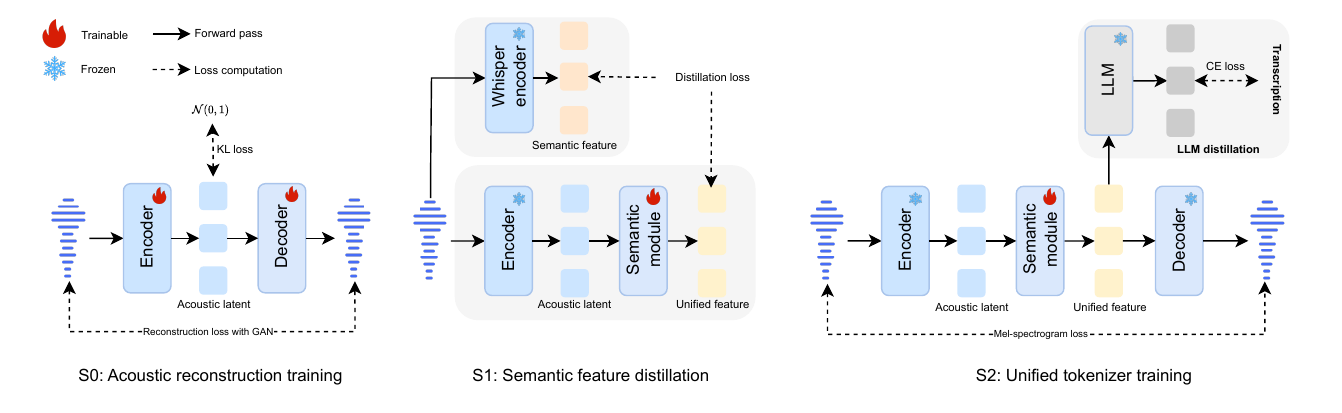}
        \vspace{-2mm}
        \caption{Three training stages. }
        \label{Ming_unitok-training}
        \vspace{-2mm}
    \end{subfigure}
    \caption{The overall framework of \mingaudiotokenizer}
\end{figure}

Our model design is guided by three core principles that aim to create a unified representation for both understanding and generation tasks.
\begin {itemize}
    \item \textbf{Continuous Representations}
    A principal challenge in developing a unified model for both understanding and generation tasks lies in the divergent nature of their preferred input representations. Understanding tasks demonstrably benefit from continuous features that better capture nuanced semantic information. Conversely, state-of-the-art speech generation models typically rely on discrete tokens, which, when combined with frameworks like flow matching and vocoders, offer an effective modeling approach.
    To construct a truly unified architecture, a single and coherent representation is imperative. Discrete tokens are suboptimal for comprehension tasks due to issues such as semantic fragmentation. Furthermore, recent work such as DitAR \citep{jia2025ditar} has demonstrated that continuous latent representations of VAE can outperform discrete tokens in generative quality. Therefore, we employ a VAE-based tokenizer that generates a continuous and unified representation for all downstream tasks.

    \item \textbf{Joint Modeling of Low- and High-Dimensional Representations}
    This unified representation ($Z_\text{uni}$) presents a core dimensionality challenge. On the one hand, comprehension tasks require high-dimensional, semantically rich inputs for the LLM. On the other hand, generative frameworks such as flow matching often face scalability challenges with high-dimensional data, making low-dimensional latents (e.g., 32 or 64 dimensions) a more tractable choice for modeling.
    To bridge this dimensionality gap, our tokenizer is designed to explicitly link the low- and high-dimensional spaces. This is achieved through a semantic module that maps the compact latent variable $Z_\text{latent}$, to its semantically richer counterpart $Z_\text{uni}$, which is then established as input for the LLM. This architecture advantageously enables all tasks to operate from a rich, high-dimensional space while harnessing the efficiency of a compact latent for generation.

    \item \textbf{Joint Optimization of Semantic Alignment and Acoustic Reconstruction}
    Our unified representation is optimized with a joint objective that simultaneously targets semantic alignment and acoustic reconstruction.
    (1) \textbf{Semantic Alignment with a pre-trained LLM}. To enrich $Z_\text{uni}$ with semantics, we use a frozen LLM as a guide. The tokenizer is trained by backpropagating gradients from a discriminative task through the LLM, which forces $Z_\text{uni}$ to align with the LLM's semantic space. (2) \textbf{Acoustic Reconstruction}. For generation tasks, it is crucial that $Z_\text{uni}$ retains sufficient low-level acoustic details. We achieve this using $Z_\text{uni}$ as the input to a reconstruction decoder, tasked with speech reconstruction. This objective of reconstruction forces $Z_\text{uni}$ to encode fine-grained acoustic details for accurate signal reconstruction. By jointly optimizing for these two objectives, $Z_\text{uni}$ becomes a single representation encoding both rich semantics for comprehension and fine-grained acoustics for generation.

\end{itemize}

\subsubsection{Tokenizer Architecture}
Our tokenizer, detailed in Figure \ref{Ming_unitok}, is a fully transformer-based architecture designed for high efficiency. It is entirely convolution-free and consists of three core modules: an encoder, a semantic module, and a decoder.
\begin {itemize}
    \item \textbf{Encoder} The encoder processes the raw audio input and encodes it in a latent space, following a strategy similar to the TS3-Codec \citep{wu2025ts3codectransformerbasedsimplestreaming}. We reshape the raw waveform into a 2D tensor, and then each frame vector is projected into the model's hidden dimension($d_\text{model}$). Subsequently, we employ a unidirectional transformer to capture contextual information. And its final hidden state is projected to a 2 $\times d_\text{latent}$ vector to obtain VAE's latent Gaussian distribution.
    \item \textbf{Semantic module} To obtain a unified representation for downstream understanding and generation training, we introduce a semantic module built upon the pre-trained Whisper large-v3 \citep{radford2023robustwhisper} encoder. We adapt this encoder by removing its convolution layers, enabling it to directly process low-dimensional acoustic latents $Z_\text{latent}$ from our VAE to high-dimensional unified features $Z_\text{uni}$. 
    \item \textbf{Decoder} The decoder reconstructs the waveform from $Z_\text{uni}$ in a two-stage process. First, a unidirectional transformer maps $Z_\text{uni}$ to an intermediate spectral representation. A subsequent synthesis head, inspired by Vocos \citep{siuzdak2023vocos}, then predicts the full complex spectrogram (magnitude and phase) for the final synthesis via an Inverse Short-Time Fourier Transform (iSTFT).
\end {itemize}

\subsection{Unified Speech Language Model}

\label{sec:method}
\begin{figure}[t]
    \centering
    \includegraphics[width=1.0\linewidth]{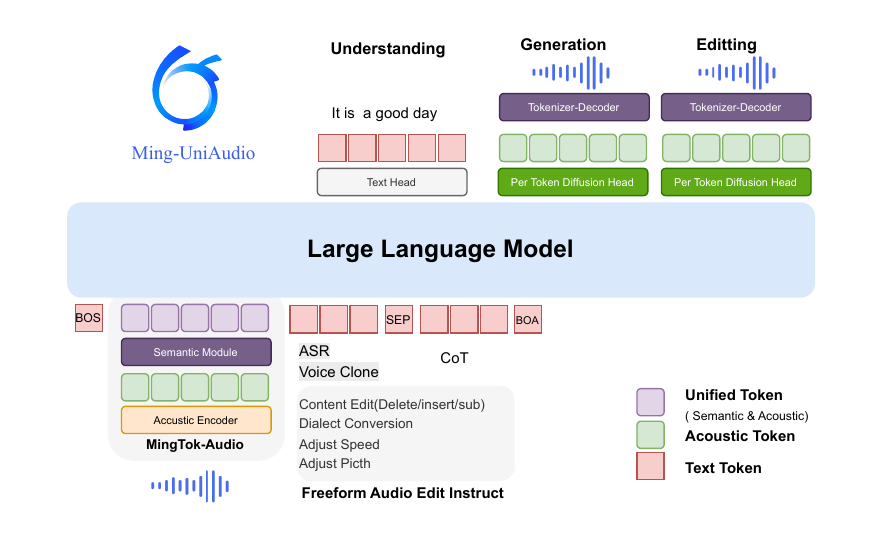 }
    \caption{Model Archieture of \model{} }
    \label{Ming_moe_uni}
    \vspace{-4mm}
\end{figure}

As illustrated in Figure \ref{Ming_moe_uni}, the core of our unified speech language model is a decoder-only LLM backbone. This architecture is designed to process a combined sequence of tokens, seamlessly integrating text tokens with the continuous unified speech tokens $Z_\text{uni}$ produced by \mingaudiotokenizer{}.   The model's versatility is enabled by a multi-head design, where different tasks are handled by specialized output modules. This unified framework allows a single model to pivot between comprehension, synthesis, and manipulation, conditioned on the specific input format and instructional prompts.

\begin{itemize}
    \item \textbf{Understanding tasks}: The LLM's final hidden states are projected by a text head to auto-regressively generate the target text sequence.
    \item  \textbf{Generative tasks}:  The hidden states are fed into a per-token diffusion head to synthesize the audio.
    \item \textbf{Editing tasks}: Leveraging this entire architecture in a "understand-then-synthesize" paradigm, the model first processes the source audio and a textual instruction to generate an intermediate representation (e.g., Chain-of-Thought), and then the per-token diffusion head synthesizes the final edited audio.
\end{itemize}


\section{Unified Tokenizer: Methodology and Training}
To effectively obtain a unified representation for both speech understanding and generation, we design a three-stage training method for the proposed architecture.

\subsection{Three-Stage Training Paradigm}
\label{sec:tokenizer-train}

\subsubsection{Acoustic reconstruction training}
In the first stage, we train an acoustic tokenizer, consisting of the encoder and the decoder, to learn a high-quality representation of the audio waveform. This process is optimized through a hybrid VAE-GAN objective, following the X-codec framework \citep{ye2024codecdoesmatterexploring}. As shown in Figure~\ref{Ming_unitok-training} (S0), the overall training objective is composed of two main loss functions, one for the discriminators($D$) and one for the generator($G$). The discriminator loss, $\mathcal{L}_D$, aims to distinguish real audio from reconstructed audio over all subdiscriminators. The generator loss, $\mathcal{L}_G$, is a composite objective designed to ensure both the accuracy of the reconstruction and the naturalness of perception. As shown in Equation~\ref{eq:loss_LG}, 

\begin{equation}
\mathcal{L}_G = \lambda_{\text{rec}} \mathcal{L}_{\text{rec}} + \lambda_{\text{adv}} \mathcal{L}_{\text{adv}} + \lambda_{\text{fm}} \mathcal{L}_{\text{fm}} + \lambda_{\text{kl}} \mathcal{L}_{\text{KL}}
\label{eq:loss_LG}
\end{equation}

where the components are: (1) a multi-scale mel-spectrogram reconstruction loss ($\mathcal{L}_\text{rec}$) to ensure perceptual audio quality; (2) an adversarial loss ($\mathcal{L}_\text{adv}$) to enhance audio naturalness by deceiving the discriminators; (3) a feature matching loss ($\mathcal{L}_\text{fm}$) on the discriminator features to stabilize GAN training; and (4) a KL-divergence loss ($\mathcal{L}_\text{KL}$) to regularize the VAE's latent space.

\subsubsection{Semantic feature distillation}
We initialize the semantic module with the weights of a pre-trained Whisper large-v3 encoder, with the help of its strong semantic priors. To adapt this module from its original mel spectrogram input to our model's acoustic latent input, we employ semantic distillation, where only the semantic module is trained, as shown in Figure~\ref{Ming_unitok-training} (S1). As defined in Equation~\ref{eq:loss_distill},  $Z_\text{semantic}$ is the output of the original frozen Whisper encoder.
\begin{equation}
    \mathcal{L}_{\text{distill}} = \text{MSE} (Z_\text{uni}, Z_\text{semantic})
    \label{eq:loss_distill}
\end{equation}

\subsubsection{Unified tokenizer training with LLM}
Following the methodology outlined in Section \ref{sec:tokenizer_design}, we perform a joint training phase for semantic alignment and speech reconstruction. During this stage, we exclusively update the parameters of the semantic module, forcing $Z_\text{uni}$ to simultaneously capture both rich semantic content and detailed acoustic features. Unlike the acoustic reconstruction training stage, the training for this task is guided only by the mel-spectrogram reconstruction loss, as we found that including GAN losses at this stage negatively impacts semantic learning.

\begin{equation}
\mathcal{L}_{\text{align}} = - \sum_{t=1}^{T} \log P(y_t | Z_\text{uni}, y_{<t})
\label{eq:align}
\end{equation}

\begin{equation}
\mathcal{L} = \lambda_{\text{align}}\mathcal{L}_{\text{align}} + \lambda_{\text{rec}}\mathcal{L}_{\text{rec}}
\label{eq:stage3}
\end{equation}

\subsection{Experimental Setup}
\subsubsection{Data \& Evaluation Metrics}
Our model is trained on a vast and diverse corpus of approximately 390,000 hours of speech data(16kHz), with a balanced 1:1 ratio of Mandarin Chinese and English. For evaluation, we report results on the public Seed-TTS-Eval \citep{anastassiou2024seedttsfamilyhighqualityversatile}\footnote{https://github.com/BytedanceSpeech/seed-tts-eval} benchmark. To comprehensively evaluate the performance of our proposed unified tokenizer, we employ several widely used objective metrics. (1) Short-Time Objective Intelligibility (\textbf{STOI}); (2) Perceptual Evaluation of Speech Quality (\textbf{PESQ}); (3) Speaker Similarity (\textbf{SPK SIM}).

\subsubsection{Training configurations}
Our unified tokenizer processes 16 kHz audio by segmenting it into non-overlapping frames of 320 samples, resulting in a 50 Hz frame rate with sliding window attention (window size=32) in all layers. Our model has a total of 1.35B parameters. We train our tokenizer following the three-stage training paradigm proposed in Section \ref{sec:tokenizer-train}. The specific configurations for each stage are detailed in Table \ref{tab:unitok-train-config}.

\begin{table}[h]
    \centering
    \small
    \begin{tabular}{cclcc}\toprule
         Training stage&  Learing rate&   Training steps&Trainable modules&  Objectives\\\midrule
         Acoustic tokenizer training&  1e-4&   200k&Encoder+ Decoder&  $\mathcal{L}_G = 15\mathcal{L}_{\text{rec}}+\mathcal{L}_{\text{adv}}+\mathcal{L}_{\text{feat}}+1e^{-4}\mathcal{L}_{\text{KL}}$\\
         Semantic feature distillation&  1e-4&   200k&Semantic module&  $\mathcal{L} = \mathcal{L}_{\text{distill}}$\\
 Unified tokenizer training& 1e-4&  200k&Semantic module& $\mathcal{L} = 2\mathcal{L}_{\text{align}} + \mathcal{L}_{\text{rec}}$\\ \bottomrule
    \end{tabular}
    \caption{Training Configurations for Unified Continuous Speech Tokenizer}
    \label{tab:unitok-train-config}
\end{table}
\subsection{Evaluation}
\subsubsection{Speech Reconstruction Performance}
The evaluation is performed according to the methods described in Section \ref{sec:tokenizer_metrics}. As shown in Table \ref{tab:audio_tokenizers}, our tokenizer demonstrates substantially superior performance in all metrics evaluated under the same conditions. This result strongly indicates that, under the continuous VAE modeling paradigm, the acoustic information of the raw speech is preserved with high quality.

\begin{table}[h]
    \small
    \centering
    \begin{tabular}{lccccccc}
        \hline
        \multirow{2}{*}{\textbf{System}} & \multirow{2}{*}{\textbf{FrameRate}} &\multicolumn{3}{c}{\textbf{SEED-ZH}}&\multicolumn{3}{c}{\textbf{SEED-EN}}\\
        \cline{3-5} \cline{6-8}& & \textbf{PESQ}↑& \textbf{SIM}↑& \textbf{STOI}↑& \textbf{PESQ}↑& \textbf{SIM}↑& \textbf{STOI}↑\\
        \hline
 \mingaudiotokenizer (ours)& 50& \textbf{4.21}& \textbf{0.96}& \textbf{0.98}&\textbf{4.04 }&\textbf{0.96 }&\textbf{0.98}\\
        MiMo-Audio-Tokenizer \cite{coreteam2025mimoaudio} & 25& 2.71 & 0.89 & 0.93 & 2.43 & 0.85 & 0.92 \\
        GLM4-Voice-Tokenizer \cite{zeng2024glm4voiceintelligenthumanlikeendtoend} & 12.5& 1.06 & 0.33 & 0.61 & 1.05 & 0.12 & 0.60 \\
        Baichuan-Audio-Tokenizer \cite{li2025baichuan} & 12.5 & 1.84 & 0.78 & 0.86 & 1.62 & 0.69 & 0.85 \\
        XY-Tokenizer \cite{gong2025xytokenizermitigatingsemanticacousticconflict} & 12.5 & 2.27 & 0.77 & 0.90 & 2.14 & 0.82 & 0.90 \\
        Mimi \cite{défossez2024moshispeechtextfoundationmodel} & 12.5 & 2.05 & 0.73 & 0.89 & 2.01 & 0.77 & 0.89 \\
        XCodec2.0 \cite{ye2025llasascalingtraintimeinferencetime} & 50 & 2.19 & 0.80 & 0.92 & 2.37 & 0.82 & 0.93 \\
        BigCodec \cite{xin2024bigcodecpushinglimitslowbitrate} & 80 & 2.26 & 0.81 & 0.92 & 2.22 & 0.80 & 0.91 \\
        \hline
    \end{tabular}
    \caption{Comparison of reconstruction performance across different acoustic tokenizers.
}
    \label{tab:audio_tokenizers}
\end{table}

\subsubsection{Downstream TTS Performance}
To validate the capabilities of our unified tokenizer for generation, we evaluated its performance on downstream TTS tasks. $Z_\text{uni}$ serves as the input to the LLM and a per-token flow matching head is employed for generative modeling. The results are presented in Table \ref{tab:tok-tts}. Our continuous feature-based TTS, built upon the unified tokenizer, significantly outperforms our previous discrete-token model, Ming-lite-omni, despite using the same training data and model scale. This result highlights the superiority of the unified representation for generative tasks.

\begin{table}[h]
\small
\centering
\begin{tabular}{@{}lllll@{}}
\toprule
\textbf{System }                                                         & \textbf{Seed-zh WER(\%)} & \textbf{Seed-zh SIM} & \textbf{Seed-en WER(\%)} & \textbf{Seed-en SIM} \\ \midrule
Seed-TTS \cite{Seed_TTS} & 1.12            & \textbf{0.80}        & 2.25            & \textbf{0.76}        \\
Ming-Omni-Lite \cite{Inclusion2025MingOmniAU} & 1.69            & 0.68        & 4.31            & 0.51        \\
\textbf{\mingaudiotokenizer}-TTS                                              & \textbf{1.04}            & 0.75        & \textbf{1.54 }           & 0.68        \\ \bottomrule
\end{tabular}
\caption{Performance of \textbf{\mingaudiotokenizer} on the downstream TTS task.}
\label{tab:tok-tts}
\end{table}


\section{Pre-training the Unified Speech Language Model}

\subsection{Methodological Insights for Unified Pretraining}

The primary objective of the pretraining stage is to enable the Speech LLM to achieve state-of-the-art performance on both understanding and generation tasks. This is of significant importance for downstream applications, such as audio editing and contextual understanding. However, unified training of understanding and generation tasks presents inconsistencies in terms of iteration steps, convergence speed, and hyperparameters. To address these training inconsistencies, we distill the following methodological insights that proved crucial for pre-training:
\begin {itemize}
    \item \textbf{Stability of Representation: } The speech tokenizer \mingaudiotokenizer{}, when independently validated on downstream understanding and generation tasks, can achieve strong performance. This indicates that the unified representation of \mingaudiotokenizer{}, denoted as $Z_{\text{uni}}$, is inherently rich in both semantic and acoustic information. However, there are disparities between the two tasks in terms of data volume, training steps, and learning rates. A stable representation is fundamental for both tasks. However, unfreezing all parameters (i.e. full fine-tuning) can easily lead to feature drift, resulting in an imbalanced performance between understanding and generation. Consequently, freezing the tokenizer and selecting a parameter-free token compression module, especially during the initial training phase, is crucial for maintaining the stability of the tokenizer's semantic and acoustic representations.

    \item \textbf{Closed-loop Flow of Representation:} Semantic and acoustic features must be able to flow in a closed loop throughout the training process. Otherwise, representation skew may occur, biasing the model's performance towards a specific task. It is crucial that in modules such as the tokenizer's semantic module, adapters of LLM, token compressor, and per-token flow matching head, both semantic and acoustic information can be freely transmitted. This prevents the representation from wavering between understanding and generation objectives at any stage, a phenomenon we term \textit{"Representation Confusion"}.
    
    \item \textbf{Joint Training:} Tasks with different convergence steps, volumes of training data, and levels of difficulty must be properly scheduled. Performance in understanding tasks is typically positively correlated with the number of tokens consumed. In contrast, the performance of speech generation, using an autoregressive model with a per-token diffusion head, is generally associated with the number of iteration steps. Therefore, to efficiently utilize computational resources and enable both understanding and generation tasks to achieve optimal performance simultaneously, a specific strategy should be adopted.
\end{itemize}


\subsection{Implementation Details}
\label{sec:pretrain_implementation}
The overall architecture employs a 16.8B-parameter Mixture-of-Experts (MoE) large language model (2.8B active) for autoregressive generation, as illustrated in Figure \ref{Ming_moe_uni}.

\paragraph{\textbf{Audio Token Compression}} is one of the most influential strategies.
After the audio tokenizer generates high-dimensional features from the input audio, there are multiple approaches to downsample the feature sequence to the desired input frame rate of the LLM.
In addition to the temporal pooling method that operates on adjacent tokens shown Equation \ref{eq:pooling}, an alternative approach (cross-attention compressor) involves employing a transformer module to aggregate information from the chunk into the [CLS] token.

\begin{equation}
\label{eq:pooling}
Z_{\text{uni}}'[i] = \mathtt{Pooling}(Z_{\text{uni}}[pi:(p+1)i]) \text{\quad for } i = 0, \dots, \left\lfloor \dfrac{l_\text{uni}}{p}\right\rfloor - 1
\end{equation}
where $l_\text{uni}$ is the length of $Z_{\text{uni}}$. To investigate the influence of the two methods, we conducted a study whose results are shown in Table \ref{tab:pretrain_ablation_performance_downsampling}. Clearly, the pooling strategy achieves significantly better results.
This suggests that the downsampling process itself does not require a highly complex mapping. Still, an efficient and accurate transmission of the original features is required to maximize the information available to downstream modules.

\begin{table}[h]
    \small
    \centering
    \resizebox{\textwidth}{!}{
    \begin{tabular}{lc@{\ }c@{\ }c@{\ }c@{\ }c@{\ }c@{\ }cccc}
        \hline
        \multirow{2}{*}{\textbf{Configuration}} & \multicolumn{7}{c}{\textbf{Und WER(\%) $\downarrow$}} & \multicolumn{3}{c}{\textbf{Gen WER(\%) $\downarrow$ | SIM$\uparrow$}} \\
        \cmidrule(lr){2-8} \cmidrule(lr){9-11}
        & \footnotesize \textbf{aishell1} & \footnotesize \textbf{aishell2-ios} & \footnotesize \textbf{Fleurs-zh} & \footnotesize \textbf{LS-clean} & \footnotesize \textbf{LS-other} & \footnotesize \textbf{Voxpopuli-en} & \footnotesize \textbf{AVG} & \footnotesize \textbf{Seed-zh} & \footnotesize \textbf{Seed-en} & \footnotesize \textbf{AVG} \\
        \hline
        Baseline (pooling) & \textbf{3.23} & \textbf{3.62} & \textbf{5.53} & \textbf{2.26} & \textbf{6.02} & \textbf{10.41} & \textbf{5.18} & \textbf{8.07} | \textbf{0.61} & \textbf{22.07} | \textbf{0.39} & \textbf{15.07} | \textbf{0.50} \\
        Baseline (cross-attention) & 5.69 & 7.05 & 6.95 & 3.64 & 10.06 & 13.97 & 7.89 & 8.88 | 0.48 & 21.89 | 0.26 & 15.39 | 0.37 \\
        \hline
    \end{tabular}
    }
    \caption{Performance comparison of the downsampling strategies. Both experiments are performed on Qwen-0.5B.}
    \label{tab:pretrain_ablation_performance_downsampling}
\end{table}

\begin{figure}[htbp]
    \centering

    \begin{subfigure}{0.48\textwidth}
        \includegraphics[width=\linewidth]{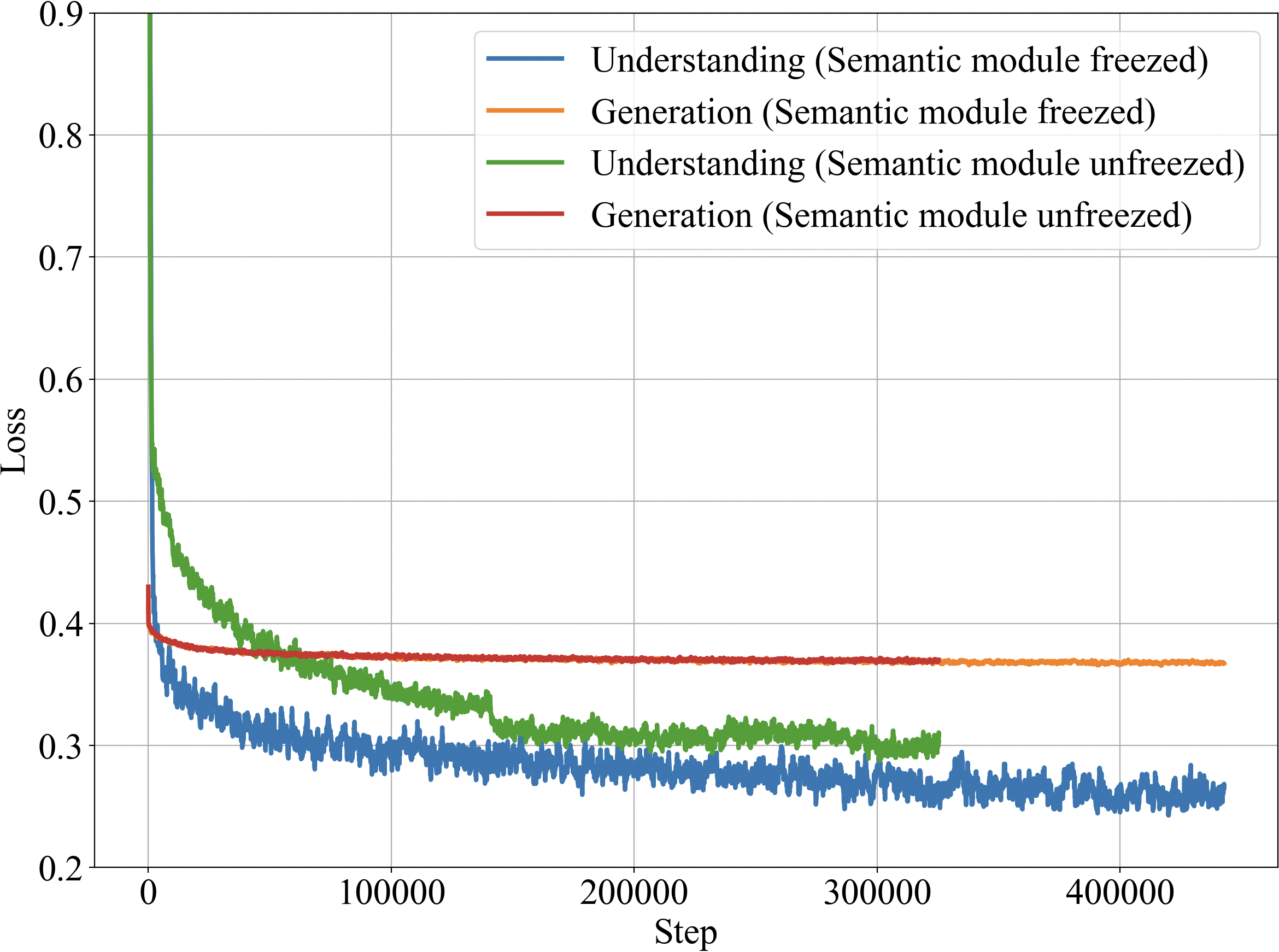}
        \caption{Semantic module freezing}
        \label{fig:pretrain_ablation_semantic}
    \end{subfigure}
    \hfill
    \begin{subfigure}{0.48\textwidth}
        \includegraphics[width=\linewidth]{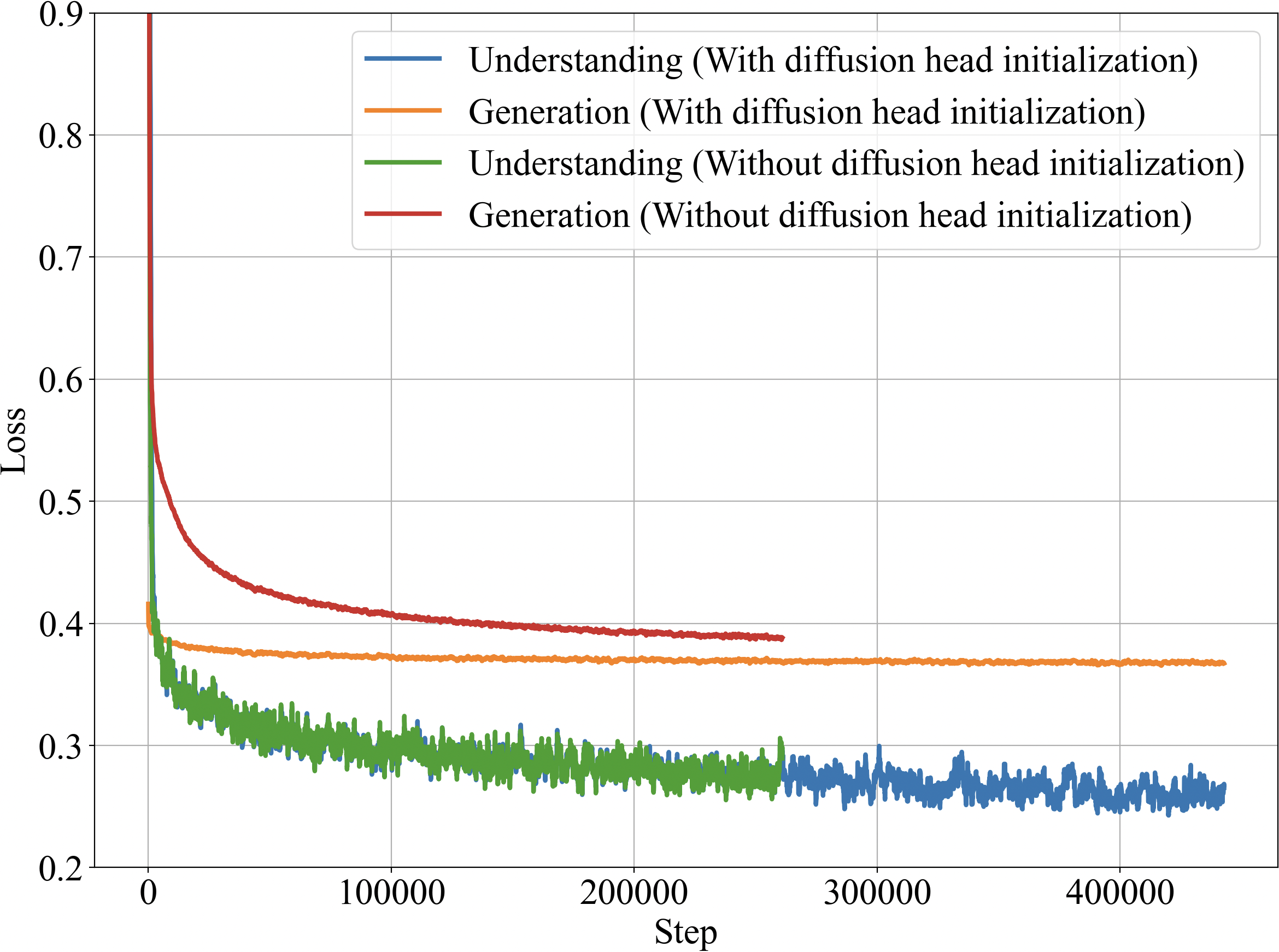}
        \caption{Diffusion head initialization}
        \label{fig:pretrain_ablation_diffusion}
    \end{subfigure}

    \caption{The loss curve of the ablation study for semantic module freezing and diffusion head initialization.}
    \label{fig:pretrain_ablation}
\end{figure}

\paragraph{\textbf{Semantic Module Freezing}} During the early stages of pre-training, the output feature distribution of the semantic module may undergo significant changes during training of understanding and generation tasks, leading to unstable convergence or slower convergence rates.
A comparative experiment on this phenomenon is presented in Table \ref{tab:pretrain_ablation_performance} and Figure \ref{fig:pretrain_ablation_semantic}.
As can be observed, when the semantic module is not frozen, the performance of both the understanding and the generation tasks deteriorates significantly, and its convergence speed is slower.
This indicates the presence of a clear representation inconsistency between understanding and generation tasks.

\paragraph{\textbf{Diffusion Head Initialization}} To accelerate the convergence of the unified speech language model, we aim to verify whether initializing the diffusion head parameters with those already trained on generation tasks can lead to a better result.
Thus, we trained a Qwen-0.5B based model in a single-task manner with the same training setup and loaded its weights before the large-scale training started.
We evaluated the performance of this strategy on Qwen-0.5B, and the results are presented in Table \ref{tab:pretrain_ablation_performance} and Figure \ref{fig:pretrain_ablation_diffusion}.
With almost no degradation in the performance of the understanding task, the performance of the generation task is significantly improved. In addition, the baseline configuration with diffusion head initialization achieves a convergence speed approximately 2.0 times that of the ablation experiments.

\begin{table}[h]
    \small
    \centering
    \resizebox{\textwidth}{!}{
    \begin{tabular}{lc@{\ }c@{\ }c@{\ }c@{\ }c@{\ }c@{\ }cccc}
        \hline
        \multirow{2}{*}{\textbf{Configuration}} & \multicolumn{7}{c}{\textbf{Und WER(\%) $\downarrow$}} & \multicolumn{3}{c}{\textbf{Gen WER(\%) $\downarrow$ | SIM$\uparrow$}} \\
        \cmidrule(lr){2-8} \cmidrule(lr){9-11}
        & \footnotesize \textbf{aishell1} & \footnotesize \textbf{aishell2-ios} & \footnotesize \textbf{Fleurs-zh} & \footnotesize \textbf{LS-clean} & \footnotesize \textbf{LS-other} & \footnotesize \textbf{Voxpopuli-en} & \footnotesize \textbf{AVG} & \footnotesize \textbf{Seed-zh} & \footnotesize \textbf{Seed-en} & \footnotesize \textbf{AVG} \\
        \hline
        Baseline & \textbf{2.29} & \textbf{3.32} & 4.26 & 2.28 & 6.04 & \textbf{7.91} & 4.35 & 4.62 | 0.58 & 8.43 | 0.39 & 6.53 | 0.49 \\
        \quad $+$ Task dist. = $1:3$ & 2.38 & 3.33 & 4.39 & 2.33 & \textbf{5.90} & 8.29 & 4.44 & \textbf{3.18} | \textbf{0.60} & \textbf{5.54} | 0.42 & \textbf{4.36} | 0.51 \\
        \quad $+$ Task dist. = $1:4$ & 2.47 & 3.43 & 4.35 & 2.40 & 6.13 & 8.13 & 4.49 & 3.20 | \textbf{0.60} & 5.75 | \textbf{0.44} & 4.48 | \textbf{0.52} \\
        \quad $-$ Semantic module freezing & 4.02 & 4.54 & 7.21 & 2.81 & 7.81 & 14.78 & 6.86 & 8.11 | \textbf{0.60} & 22.49 | 0.39 & 15.30 | 0.50 \\
        \quad $-$ Diffusion head init & 2.36 & 3.33 & \textbf{4.22} & \textbf{2.27} & 5.92 & 7.92 & \textbf{4.34} & 10.02 | 0.51 & 12.64 | 0.32 & 11.33 | 0.42 \\
        \midrule
        Baseline 420k steps & 1.62 & 3.08 & 4.33 & 1.76 & 4.91 & 8.46 & 4.03 & 2.27 | 0.63 & 6.59 | 0.44 & 4.43 | 0.54 \\
        \quad $+$ Annealing & \textbf{1.47} & \textbf{2.82} & \textbf{4.06} & \textbf{1.59} & \textbf{4.39} & \textbf{8.37} & \textbf{3.78} & \textbf{1.97} | \textbf{0.64} & \textbf{6.05} | \textbf{0.46} & \textbf{4.01} | \textbf{0.55} \\
        \hline
    \end{tabular}
    }
    \caption{Performance comparison of several training strategies.}
    \label{tab:pretrain_ablation_performance}
\end{table}

\paragraph{\textbf{Stopping Criterion For Generation with Continuous Token}}
Unlike the discrete tokenizer, which benefits from a special <EOS> token, our model requires an additional stop detector. We append a linear layer to the LLM that performs binary classification at each frame. To avoid a strong dependence on tail annotation, we propose a weakly supervised training strategy. (1) The last frame of each utterance is labeled as the positive sample(stop); (2) Then the previous three frames(300ms) of the last frame were ignored to minimize the impact of VAD errors; (3) Among the remaining frames, one frame with the highest stop score was selected as the negative sample. This method keeps a balance of positive to negative samples and mines the hardest negative samples online. The stop head is trained with a cross‐entropy loss scaled by a factor of 0.01.

\label{sec:data}

\subsection{Data Curation and Processing}

The performance of our model relies on a large-scale, diverse training dataset. We compiled our data from two primary sources: publicly available audio datasets and an in-house collection comprising web-crawled recordings and synthetically generated audio.

To ensure data quality and relevance, we implemented a multi-stage preprocessing pipeline. First, a rigorous cleaning process is applied to all data, particularly to the web-crawled audio which exhibits significant quality variance. Second, recognizing that different tasks such as understanding and generation have distinct requirements, we perform task-specific filtering. This tailored approach allows us to create high-quality, annotated data subsets optimized for a wide range of speech-related tasks, forming a solid foundation for training our unified model.

\begin{figure}[bpth]
    \centering
    \includegraphics[width=1.0\linewidth]{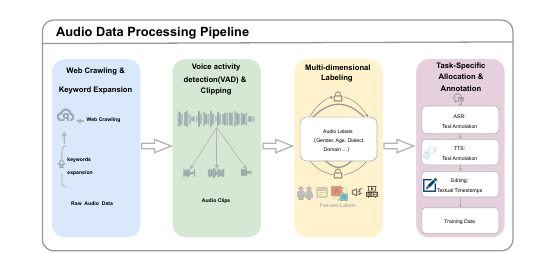}
    \caption{
    Overview of Audio Data Processing Pipeline.
    }
    \label{fig-data-overall}
\end{figure}

\subsection{Training}

\paragraph{\textbf{Large-scale training stage}} This stage aims to provide the LLM with enough large-scale data to establish its fundamental alignment capability.
During this phase, the model was trained for 200,000 steps, with a 1:3 training step ratio between understanding and generation tasks.
Throughout the entire training process, the parameters of the semantic module within the audio tokenizer were kept frozen.

\paragraph{\textbf{Annealing stage}} The purpose of this stage is to perform annealing using higher-quality data and a smaller learning rate, thereby stabilizing the model parameters while further improving the overall performance.
All other training configurations remain the same as in the previous stage.

\paragraph{\textbf{Full fine-tuning stage}} This stage continues training with the unfrozen semantic module parameters, building on the previous stage.
At the same time, the proportion of generation task training is moderately increased, with the aim of further improving the performance of the generation task while minimally affecting the performance of the understanding task.

\begin{table}[h]
    \small
    \centering
    \begin{tabular}{lcccc}
        \toprule
        \multirow{2}{*}{\textbf{Hyper-parameter}} & \multicolumn{3}{c}{\textbf{Pre-training}}  \\
        \cmidrule(lr){2-4}
        & \textbf{Large-scale training} & \textbf{Annealing} & \textbf{Full fine-tuning} \\
        \midrule
        LR (Semantic Module)     & frozen & frozen & 1e-5 \\
        LR (Others)              & 1e-4 & 1e-5 & 1e-5 \\
        LR scheduler             & constant & constant & constant \\
        batch size (Und)         & 5500 tokens & 5500 tokens & 5500 tokens \\
        batch size (Gen)         & 2000 tokens & 2000 tokens & 2000 tokens \\
        step ratio (Und:Gen)     & 1:3 & 1:3 & 1:6 \\
        warmup steps             & 3000 & 0 & 0 \\
        AdamW eps                & 1e-5 & 1e-5 & 1e-8 \\
        Training data duration(h) (Und:Gen) & 400k:800k & 370k:370k & 370k:370k \\
        \hline
    \end{tabular}
    \caption{Training configuration across different stages. LR stands for learning rate. Und stands for understanding. Gen stands for generation.}
    \label{tab:training_configuration}
\end{table}

%

\section{Instruction-based Free-Form Speech Editing}
\subsection{Unified Paradigm for Instruction-Based Editing}

\label{sec:sft_overview}
\begin{figure}[t]
    \centering
    \includegraphics[width=0.9\linewidth]{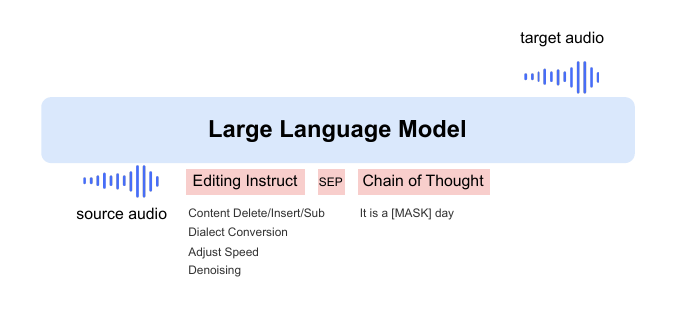}
    \caption{Comparison of \model{} with Open-Source Speech LLMs across Understanding, Generation, and Editing Tasks. }
\label{fig:sft_edit}
    \vspace{-4mm}
\end{figure}

Speech editing is a task that places high demands on both understanding and generation capabilities simultaneously. Leveraging the strong foundational capabilities of the \model{} model in both understanding and generation, we explore the training paradigms, data construction methods, and evaluation metrics for free-form instruction-based speech editing. 
The core challenge in unifying speech understanding and generation stems from a fundamental inconsistency in representation. Understanding tasks thrive on continuous, semantically rich features, while state-of-the-art generation models often rely on discrete tokens. Although effective for synthesis, these discrete tokens sacrifice fine-grained acoustic information, which in turn degrades performance on understanding tasks. This forces a fragmented approach: using continuous tokens for understanding and discrete ones for generation. The result is two disparate tokenization schemes that fundamentally prevent the development of a truly unified, high-quality representation.

Building upon a pre-trained, continuous and unified tokenizer \mingaudiotokenizer{}, as well as a pre-trained Speech LLM \model{} that unifies both understanding and generation tasks, our approach enables high-quality speech understanding and speech generation using a single, unified representation. This architecture perfectly aligns with the "understand-then-synthesize" paradigm required for speech editing tasks as shown in Figure \ref{fig:sft_edit}. Ultimately, our model supports a variety of editing tasks. The corresponding input and output formats for each task are detailed in Table \ref{tab:tasks}. To the best of our knowledge, this is the first speech editing model capable of following human instructions without requiring any additional operations such as timestamp alignment.

\begin{table}[h!]
\centering
\small
\begin{tabular}{l@{\hspace{0.8em}}lll@{}}
\toprule 
 Type&\textbf{Task} & \textbf{Input} & \textbf{Output} \\
\midrule 

 &Deletion &  SourceSpeech, delete the word `hello' in...& $\mathtt{CoT_{text}}$, TargetSpeech \\
 Semantic&Insertion &  SourceSpeech, insert the word `hello' in...& $\mathtt{CoT_{text}}$, TargetSpeech \\
 &Substitution &  SourceSpeech, substitute the word `hello' with `world'& $\mathtt{CoT_{text}}$, TargetSpeech \\
\midrule 
 &Denoise &  SourceSpeech, denoise the audio& TargetSpeech \\
 &Speed Alteration &  SourceSpeech, adjust the speed to 1.5& TargetSpeech \\
 Acoustic&Pitch Alteration &  SourceSpeech, shift the pitch by 3 steps& TargetSpeech \\
 &Volume Alteration &  SourceSpeech, adjust the volume to 1.5& TargetSpeech \\
 &Dialect Conversion &  SourceSpeech, change the accent of the speech to Chengdu& TargetSpeech \\
\bottomrule 
\end{tabular}
\caption{A summary of various speech editing tasks and their task formulation. }
\label{tab:tasks}
\end{table}

We designed the training scheme for free-form instruction-based editing based on \textit{following principles}:
\begin{itemize}
    \item \textbf{Global vs. Local:} Semantic and acoustic editing tasks exhibit a fundamental local versus global distinction. Semantic editing is typically local: it modifies content within a specific region while requiring the surrounding audio (content, timbre, prosody) to remain unchanged. The edited segment must also maintain acoustic consistency with its context. In contrast, acoustic editing (e.g., accent conversion, denoising) is often global, altering acoustic features across the entire utterance while preserving the original text. This distinction requires different modeling approaches.
    \item \textbf{Locate then Modify:}  Given their local nature, semantic editing tasks require a two-stage process: first locating the target segment and then performing the modification, which we formulate as a Chain of Thought for Semantic Tasks.
    \item \textbf{Joint Training of Complementary Tasks:} Many training tasks can be viewed as complementary pairs, such as denoising versus adding background music, and text addition versus deletion. Consequently, multitask joint training is expected to enhance overall model performance by leveraging these complementary relationships.
\end{itemize}


Based on these principles, we propose a unified generative model that seamlessly integrates semantic understanding and acoustic synthesis. To address the "Global vs. Local" distinction, our model operates on the unified representation ($Z_\text{uni}$) for content manipulation and uses a per token diffusion head for high-quality generation, accommodating both local semantic edits and global acoustic changes. For the "Locate then Modify" principle in semantic tasks, we formulate the process as an instruction-guided sequence-to-sequence transformation: the model first autoregressively generates a target semantic sequence, effectively performing a "chain-of-thought" reasoning to determine the edited content and its location.

\paragraph{\textbf{Acoustic Tasks}}
We have designed a unified paradigm for a wide array of acoustic editing tasks. These include paralinguistic modifications like emotion and dialect conversion, physical attribute adjustments such as speed, pitch, and volume, and other tasks like denoising and adding background sound. This framework takes an input audio and an instruction to produce an edited audio output directly. Crucially, it avoids generating any textual or other semantic outputs, which allows it to accommodate diverse types of output audio flexibly. This paradigm demonstrates powerful robustness and performs exceptionally well across various content-preserving editing tasks, where the underlying textual content remains unchanged.

\paragraph{\textbf{Semantic Tasks}}
For semantic editing tasks, we adopt an instruction-following format allowing us to effectively leverage the pre-trained model's ability to comprehend and execute instructions. The data is structured as follows:
\[
    \mathtt{SourceSpeech}_{z_{\mathtt{uni}}} + \mathtt{Instruction}_{\mathtt{text}}  \rightarrow \mathtt{CoT}_{\mathtt{text}} + \mathtt{TargetSpeech}_{z_{\mathtt{latent}}}
\]
We introduced several key innovations to achieve free-form instruction following:
\begin{itemize}

\item \textbf{CoT for Textual Guidance:} We incorporate a chain-of-thought (CoT) \citep{wei2023chainofthoughtpromptingelicitsreasoning} step where the model first generates the target text of the edited speech. This intermediate textual output significantly improves the pronunciation accuracy and instruction-following capabilities of the final synthesized audio.

\item  \textbf{Explicit Edit Localization:} Inserting a special [MASK] token into the CoT text explicitly demarcates the edit region. This provides the model with precise positional cues, guiding its focus to the intended segment for modification and enhancing its positional awareness.

\item  \textbf{Weighted Loss for Edited Regions:} To prioritize learning the complex synthesis of new content, a higher loss weight is applied to the edited audio segments. This strategy distinguishes these regions from non-edited ones, where reconstruction can be simplified by copying from the source audio.

\end{itemize}

\begin{figure}
    \centering
    \includegraphics[width=1\linewidth]{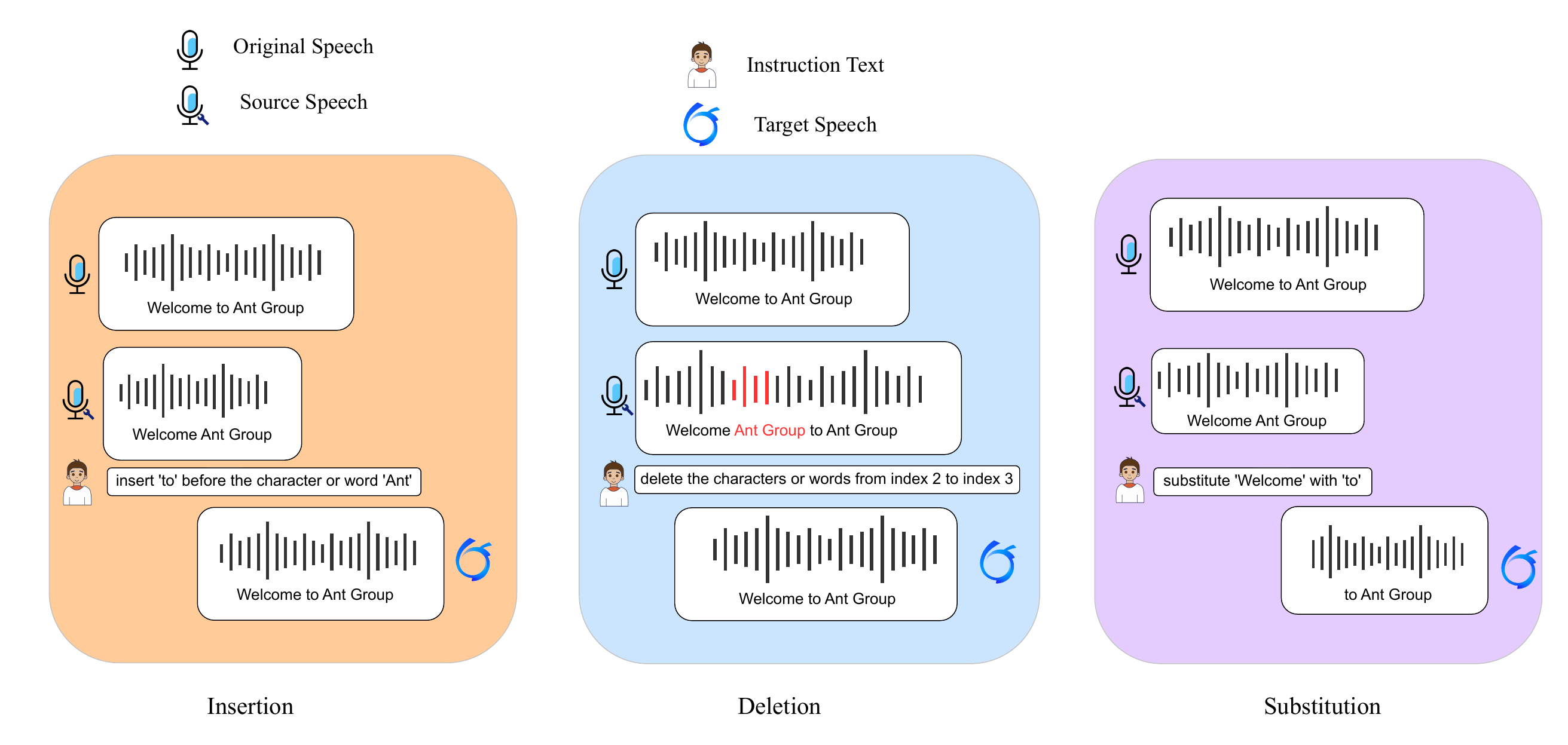}
    \caption{Speech Edit Data Construction Method}
    \label{fig:speech_edit_data}
\end{figure}

\subsection{Instruct Tuning Data}
\label {sec:instruction data}

We construct speech editing training dataset by the following methods as shown in Figure \ref{fig:speech_edit_data}:

\textbf{Insertion}: This task involves inserting content at a specified position, which can be defined by a text index or as a location relative to a specific word (i.e., before or after it). Training pairs are constructed with an audio segment that contains a removed portion as input and the original complete audio as the target.
\newline
\textbf{Deletion}: As a dual of insertion, this task aims to remove specified content. We synthesize the input by inserting a random audio artifact into a clean recording. The target is the original, unmodified audio, thus training the model to remove the artifact.
\newline
\textbf{Substitution}: This task aims to replace a segment of audio with different content. We construct the input by relocating a segment of the original audio to overwrite a different, non-overlapping segment within the same recording. The target is the original, unmodified audio, training the model to restore the overwritten section.
\newline
\textbf{Denoising}: Noisy inputs are generated by augmenting clean audio with noise from the MUSAN \citep{snyder2015musanmusicspeechnoise} dataset. The target is the original, clean audio.
\newline
\textbf{Speed and Pitch Alteration}: We processed the audio using open source tools to alter its speed and pitch. The processed audio served as the training target, with the original audio and an instruction as input.
\newline
\textbf{Dialect Conversion}: We built a Mandarin-dialect parallel corpus. To enrich speaker diversity, we further apply voice conversion (VC) to generate over 1,000 distinct timbres for the training data.

\subsection{Freeform-Audio-Edit Benchmark}
We construct \editbenchmark{}, the first universal free-form speech editing benchmark for semantic and acoustic tasks without timestamp condition, including speech content insertion/deletion/substitution, emotion/accent conversion, speed/pitch/volume alteration, etc.

\subsubsection{Semantic Test Sets}
Public benchmarks for the semantic speech editing task are scarce, which hinders the direct comparison of different methods. In addition, existing evaluation sets are predominantly designed for region-based models. To our knowledge, \editbenchmark{} is the first benchmark constructed for free-form speech editing models.

\textbf{Dataset Construction:} We select 896 Chinese and 655 English samples from the seed-tts test set. For each audio clip, it is first randomly assigned to one of three tasks: deletion, insertion, or substitution. Next, we obtain its transcription and leverage LLM to generate a natural language instruction for the assigned task and the corresponding edited text. Finally, the edited text generated by the LLM is manually verified to ensure that it aligns with the expected outcome of the given instruction. 

\textbf{Dataset statistics:} The instructions for deletion, insertion, and substitution tasks can be broadly categorized into two types: \textit{index-based}, which specifies an operation on content from positions $i$ to $j$, and \textit{content-based}, which targets specific characters or words for editing. 

In our initial version, the editing instructions are written in English for both Chinese and English audio, and their phrasing was relatively limited in variety (see Appendix \ref{sec:ap_exampels} for examples of \editbenchmark{}). We refer to this as the \textit{basic} version; the dataset statistics for this \textit{basic} version are shown in Table \ref{tab:s_benchmark_basic}.

\begin{table}[h]
    \small
    \centering
    \begin{tabular}{lcccccc}
        \hline
        \multirow{2}{*}{\textbf{Instruction Type}} & \multicolumn{3}{c}{\textbf{Zh}} & \multicolumn{3}{c}{\textbf{En}} \\
        \cmidrule(lr){2-4} \cmidrule(lr){5-7}
         & \textbf{deletion} & \textbf{insertion} & \textbf{substitution} & \textbf{deletion} & \textbf{insertion} & \textbf{substitution} \\
        \hline
        Index-based & 92 & 65 & 29 & 47 & 79 & 29 \\
        Content-based & 78 & 105 & 130 & 133 & 81 & 150 \\
        Total & 170  & 170 & 159  & 180 & 160 & 179 \\
        \hline
    \end{tabular}
    \caption{Dataset statistics for \textbf{\textit{basic}} semantic test sets of \editbenchmark{}.}
    \label{tab:s_benchmark_basic}
\end{table}
To better align the editing instructions with the real user expressions, we construct a second version of those test sets, referred to as the \textit{full} version. In addition to increasing the number of test examples per task, we deliberately emphasize the naturalness and diversity of the instructions when generating them with the LLM. Furthermore, our preliminary experiments revealed that the three tasks vary in difficulty, with substitution proving to be the most challenging. Consequently, we increase the proportion of substitution tasks in our test set to ensure a more rigorous evaluation. Detailed statistics of the \textit{full} version dataset are provided in Table \ref{tab:freeform_audio_edit_data}.

\begin{table}[h]
    \small
    \centering
    \begin{tabular}{lcccccc}
        \hline
        \multirow{2}{*}{\textbf{Instruction Type}} & \multicolumn{3}{c}{\textbf{Zh}} & \multicolumn{3}{c}{\textbf{En}} \\
        \cmidrule(lr){2-4} \cmidrule(lr){5-7}
         & \textbf{deletion} & \textbf{insertion} & \textbf{substitution} & \textbf{deletion} & \textbf{insertion} & \textbf{substitution} \\
        \hline
        Index-based & 186 & 180 & 36 & 138 & 100 & 67 \\
        Content-based & 95 & 110 & 289 & 62 & 99 & 189 \\
        Total & 281 & 290 & 325 & 200 & 199 & 256 \\
        \hline
    \end{tabular}
    \caption{Dataset statistics for \textbf{\textit{full}} semantic test sets of \editbenchmark{}.}
    \label{tab:freeform_audio_edit_data}
\end{table}

\subsubsection{Acoustic Test Sets}
\textbf{Dataset Construction:} The construction procedure for the acoustic test sets is similar to that of the semantic test sets: the audio are sourced from the seed-tts test set, and a LLM generates task-specific editing instructions. Unlike the semantic test sets, no text labels are required, eliminating the manual verification step.

\textbf{Dataset statistics:} Per-task test sample counts for the acoustic test sets in \editbenchmark{} are presented in Table \ref{tab:a_benchmark}.
\begin{table}[h]
    \small
    \centering
    \begin{tabular}{lcc}
        \hline
        \textbf{Task Type} & {\textbf{Zh}} & {\textbf{En}} \\
        \hline
        Emotion conversion & 84 & 72  \\
        Speed alteration & 50 & 50  \\
        Pitch alteration & 50 & 50 \\
        Volume alteration & 50 & 50 \\
        Dialect conversion & 250 & -- \\
        \hline
    \end{tabular}
    \caption{Dataset statistics for acoustic test sets of \editbenchmark{}.}
    \label{tab:a_benchmark}
\end{table}
\subsubsection{Evaluation Metrics}

For semantic editing task, we report the WER and SIM to respectively assess the intelligibility of speech and speaker similarity following seed-tts \citep{Seed_TTS}. In addition, we define ACC and no-edit WER to measure the audio editing accuracy within the editing area and the audio WER outside the editing area.
For acoustic editing tasks, the evaluation metrics vary depending on the specific task. Specifically, for the speech denoising task, we use the DNSMOS \citep{9746108} to evaluate the quality of the enhanced speech. This score ranges from 0 to 5, where a higher score indicates better audio quality. For the remaining tasks, we primarily employ WER and SIM as the evaluation criteria. We also report the relative duration error (RDE) for speed alteration and the relative amplitude error (RAE) for volume alteration; lower values indicate closer adherence to the target specifications.

\section{Results and Analysis}

\subsection{Understanding}

As presented in Table \ref{tab: main_asr_perf}, \model{} achieves performance comparable to leading open-source models on several public benchmarks. Notably, due to its specialized multi-dialect training, it significantly outperforms existing models on our in-house dialect test set.

Furthermore, the model demonstrates a strong capability in leveraging contextual information, as evidenced by its performance on the ContextASR Bench \citep{Wang2025ContextASRBenchAM}. According to the results in Table \ref{tab: main_context_asr_perf}, \model{} achieves SOTA performance on 8 of the 12 subtasks. This suggests a strong potential for robust performance in real-world scenarios such as multi-turn dialogue and hotword enhancement.

\begin{table*}[h!]
\centering
\resizebox{\textwidth}{!}{%
\begin{tabular}{@{}llcccccccccccccccccccccc@{}}
\toprule
\textbf{Datasets} & \textbf{Model} & \multicolumn{7}{c}{\textbf{Performance}} \\
\midrule

& & \multicolumn{1}{c}{\textbf{aishell2-ios}} & \multicolumn{1}{c}{\textbf{LS-clean}}  & \multicolumn{1}{c}{\textbf{Hunan}}  & \multicolumn{1}{c}{\textbf{Guangyue}}  & \multicolumn{1}{c}{\textbf{Chuanyu}}  & \multicolumn{1}{c}{\textbf{Shanghai}} \\
\cmidrule(lr){3-3} \cmidrule(lr){4-4} \cmidrule(lr){5-5} \cmidrule(lr){6-6} \cmidrule(lr){7-7} \cmidrule(lr){8-8} \cmidrule(lr){9-9}
\multirow{4}{*}{\makecell[l]{\textbf{Understanding} \\ \textbf{ASR}}}
& Kimi-Audio \cite{KimiTeam2025KimiAudioTR}        & \textbf{2.56} & \textbf{1.28} & 31.93 &  41.49 & 6.69 & 60.64 \\
& Qwen2.5 Omni \cite{xu2025qwen2}  & 2.75 & 1.80 & 29.31 &  10.39 & 7.61 & 32.05 \\
& Qwen2 Audio \cite{chu2023qwenaudio} & 2.92 & 1.60 & 25.88  & 7.59 & 7.77 & 31.73 \\
& \modelraw         &   2.84   &  1.62    & \textbf{9.80}  & \textbf{5.51} & \textbf{5.46} & \textbf{14.65} \\

\bottomrule
\end{tabular}
} 
\caption{Comparison of ASR performance on various audio benchmark datasets. The best results are in \textbf{bold}.}
\label{tab: main_asr_perf}
\end{table*}

\begin{table*}[h!]
\centering
\resizebox{\textwidth}{!}{%
\begin{tabular}{@{}llcccccccccccccccccccccc@{}}
\toprule
\textbf{Datasets} & \textbf{Model} & \multicolumn{4}{c}{\textbf{Performance}} \\
\midrule

& & \multicolumn{1}{c}{\textbf{Speech-English}} & \multicolumn{1}{c}{\textbf{Dialogue-English}} & \multicolumn{1}{c}{\textbf{Speech-Mandarin}} & \multicolumn{1}{c}{\textbf{Dialogue-Mandarin}} \\
& & \multicolumn{1}{c}{\footnotesize \textbf{WER | NE-WER | NE-FNR}} & \multicolumn{1}{c}{\footnotesize \textbf{WER | NE-WER | NE-FNR}} & \multicolumn{1}{c}{\footnotesize \textbf{WER | NE-WER | NE-FNR}} & \multicolumn{1}{c}{\footnotesize \textbf{WER | NE-WER | NE-FNR}} \\
\cmidrule(lr){3-3} \cmidrule(lr){4-4} \cmidrule(lr){5-5} \cmidrule(lr){6-6}
\multirow{7}{*}{\makecell[l]{\textbf{Understanding} \\ \textbf{Context ASR}\\\cite{wang2025contextasrbenchmassivecontextualspeech}}} 
& Qwen2-Audio \cite{chu2023qwenaudio} &  11.49 | 27.27 | 35.08 & 13.99 | 33.02 | 32.92 & 9.92 | 24.10 | 30.02 & 7.00 | 22.76 | 26.17 \\
& Baichuan-Audio \cite{li2025baichuanaudiounifiedframeworkendtoend} & 7.52 | 5.87 | 4.55 & 5.66 | 10.01 | 3.64 & 2.16 | 6.65 | \textbf{2.35} & 2.96 | 11.48 | 3.94 \\
& Kimi-Audio \cite{KimiTeam2025KimiAudioTR}    & \textbf{2.90} | 6.68 | 8.01 & \textbf{4.67} | 13.50 | 11.31 & 1.95 | 11.13 | 15.28 & 2.90 | 15.91 | 16.68 \\
& Baichuan-Omni-1.5 \cite{li2025baichuanomni15technicalreport} & 8.16 | 7.69 | 6.53 & 9.91 | 14.40 | 5.54 &  2.98 | 8.39 | 4.71 & 5.00 | 16.83 | 7.84 \\
& Qwen2.5-Omni-3B  \cite{xu2025qwen2} & 3.99 | 7.80 | 9.69 & 4.83 | 14.36 | 12.85 & 2.13 | 10.55 | 14.11 & 3.12 | 15.07 | 15.17 \\
& Qwen2.5-Omni-7B \cite{xu2025qwen2} & 3.96 | 7.38 | 8.72 & 5.32 | 11.83 | 9.24 & 1.84 | 9.80 | 12.19 & \textbf{2.40} | 14.06 | 13.17 \\
& \modelraw         &   4.00 | \textbf{3.56} | \textbf{3.69}   &  5.34 | \textbf{8.73} | \textbf{2.53} & \textbf{1.58} | \textbf{5.98} | 2.40 & 3.04 | \textbf{9.50} | \textbf{1.48}    \\

\bottomrule
\end{tabular}
} 
\caption{Comparison of Contextual ASR performance on ContextASR Bench \citep{Wang2025ContextASRBenchAM}. The best results are in \textbf{bold}.}
\label{tab: main_context_asr_perf}
\end{table*}

\subsection{Generation}

As shown in Table \ref{tab:main_results_gen}, \model{} demonstrates strong overall performance in speech generation. For speech intelligibility, measured by Word Error Rate (WER), it achieves the best result on Chinese among the compared models (0.95) and remains highly competitive on English. While there is room for further improvement in timbre similarity (SIM), the model consistently produces highly intelligible speech, which is a primary focus of our work.

\begin{table*}[h!]
\centering

\resizebox{\textwidth}{!}{
\begin{tabular}{@{}llcccc@{}}
\toprule
\textbf{Datasets} & \textbf{Model} & \multicolumn{4}{c}{\textbf{Performance}} \\
\midrule

& & \textbf{Seed-zh WER(\%)} & \textbf{Seed-zh SIM} & \textbf{Seed-en WER(\%)} & \textbf{Seed-en SIM} \\
\cmidrule(lr){3-3} \cmidrule(lr){4-4} \cmidrule(lr){5-5} \cmidrule(lr){6-6}
\multirow{5}{*}{\textbf{Generation}}
& Seed-TTS \cite{Seed_TTS}          & 1.12 & \textbf{0.80} & 2.25 & \textbf{0.76} \\
& FireRedTTS \cite{FireRedTTS}          & 1.51 & 0.65 & 3.82 & 0.53 \\
& FireRedTTS-2 \cite{Xie2025FireRedTTS2TL}          & 1.14 & 0.736 & 1.95 & 0.65 \\
& DiTAR \cite{jia2025ditar}          & 1.02 & 0.753 & 1.69 & 0.74 \\
& F5-TTS \cite{F5_TTS}          & 1.56 & 0.74 & 1.83 & 0.65 \\
& CosyVoice 2 \cite{CosyVoice2}          & 1.45 & 0.75 & 2.57 & 0.65 \\
& CosyVoice 3-1.5B \cite{Du2025CosyVoice3T}          & 1.12 & 0.78 & 2.21 & 0.72 \\
& MiMo-Audio \cite{coreteam2025mimoaudio}        & 1.96 &  -   & 5.37 &  -   \\
& Qwen2.5-Omni-7B$_{RL}$ \cite{xu2025qwen2}  & 1.42& 0.75& 2.33&0.64\\
& Qwen3-Omni-30B-A3B-Instruct \cite{Xu2025Qwen3OmniTR} & 1.07 & - & \textbf{1.39} & - \\

& \modelraw    &  \textbf{0.95}   &  0.70    &  1.85    &  0.58          \\

\bottomrule
\end{tabular}
}
\caption{Performance comparison on speech generation benchmark datasets. The best results are in \textbf{bold}.}
\label{tab:main_results_gen}
\end{table*}

\subsection{Editting}

Table \ref{tab:main_results_edit} summarizes the objective results for our editing tasks. Overall, \model{} demonstrates robust and versatile editing capabilities across a wide range of semantic and acoustic manipulations.

\textbf{Semantic Editing}: A key feature of \modeledit{} is its ability to perform free-form semantic editing based on natural language instructions. As most existing open-source models lack this capability, direct comparisons are limited. Our internal analysis, however, reveals several key performance trends:

\begin{itemize}
    \item \textbf{Strong Task Performance}: It achieves high accuracy on core tasks, including 79.31\% for insertion (WER=3.89) and 76.62\% for substitution (WER=4.56).
    \item \textbf{Robust Generalization}: The model shows robust generalization to varying instruction complexities, evidenced by the consistent performance between the \textit{base} and \textit{full} versions of \editbenchmark{}.
    \item \textbf{Language and Task Variation}: Performance on Chinese tasks consistently surpasses that of English, likely due to a larger, higher-quality training set. Among the semantic tasks, insertion yields the lowest WER, benefiting from more coherent training targets.
\end{itemize}

\textbf{Acoustic Editing}: It excels in voice-conversion tasks, notably achieving high stylistic consistency and semantic stability in dialect conversion, alongside strong results in volume and speed alterations. In contrast, pitch alteration presents a challenge to speaker similarity, likely due to target data quality. On DNS Challenge test set \citep{reddy20_interspeech}, its DNSMOS score (3.26 OVRL) is highly competitive with general-purpose models and specialized approaches, validating our unified approach.

\begin{table*}[h]
\centering
\resizebox{\textwidth}{!}{%
\begin{tabular}{@{}llcccc@{}}
\toprule
\textbf{Datasets} & \textbf{Model} & \multicolumn{4}{c}{\textbf{Performance}} \\
\midrule

& & \textbf{WER(\%) zh | en} & \textbf{ACC zh | en} & \textbf{SIM zh | en} & \textbf{no-edit WER(\%) zh | en} \\
\cmidrule(lr){3-3} \cmidrule(lr){4-4} \cmidrule(lr){5-5} \cmidrule(lr){6-6}
{\textbf{Deletion-basic}}
& \multirow{2}{*}{\modelraw}         &   11.89 | 14.85   &   100 | 82.22   &  0.78 | 0.76    &   11.49 | 24.26   \\
{\textbf{Deletion-full}} &&   22.92 | 27.60   &   82.92 | 85   &  0.81 | 0.74    &   17.50 | 35.21   \\
\midrule

& & \textbf{WER(\%) zh | en} & \textbf{ACC zh | en} & \textbf{SIM zh | en} & \textbf{no-edit WER(\%) zh | en} \\
\cmidrule(lr){3-3} \cmidrule(lr){4-4} \cmidrule(lr){5-5} \cmidrule(lr){6-6}
{\textbf{Insertion-basic}}
& \multirow{2}{*}{\modelraw}         &   3.42 | 6.63   &   80 | 71.43   &  0.83 | 0.79    &   3.52 | 17.70   \\
{\textbf{Insertion-full}}
&        &   3.89 | 7.592   &   79.31 | 62.31   &  0.83 | 0.79    &   4.10 | 18.84   \\
\midrule

& & \textbf{WER(\%) zh | en} & \textbf{ACC zh | en} & \textbf{SIM zh | en} & \textbf{no-edit WER(\%) zh | en} \\
\cmidrule(lr){3-3} \cmidrule(lr){4-4} \cmidrule(lr){5-5} \cmidrule(lr){6-6}
{\textbf{Substitution-basic}}
& \multirow{2}{*}{\modelraw}          &   4.52 | 8.99   &   78.62 | 59.78   &   0.82 | 0.78   &   4.63 | 19.28   \\
{\textbf{Substitution-full}}
&         &   4.56 | 7.64   &   76.62 | 65.62   &   0.83 | 0.77   &   4.75 | 18.39   \\
\midrule

& & \textbf{WER(\%)} & \textbf{ACC} & \textbf{SIM} \\
\cmidrule(lr){3-3} \cmidrule(lr){4-4} \cmidrule(lr){5-5}
\multirow{2}{*}{\makecell[l]{\textbf{Dialect} \\ \textbf{conversion}}}
& \modelraw         &   8.93   &   0.50  &   0.66  &      \\
&                   &      &      &      &      \\
\midrule


& & \textbf{WER(\%) zh | en} & \textbf{SIM zh | en} & \textbf{RDE(\%) zh | en} &\\
\cmidrule(lr){3-3} \cmidrule(lr){4-4} \cmidrule(lr){5-5}
\multirow{1}{*}{\textbf{Speed Alteration}}
& \modelraw         &   5.88 | 17.53   &   0.66 | 0.57   &   6.36 | 5.92   &      \\
\midrule

& & \textbf{WER(\%) zh | en} & \textbf{SIM zh | en} &&\\
\cmidrule(lr){3-3} \cmidrule(lr){4-4} 
\multirow{1}{*}{\textbf{Pitch Alteration}}
& \modelraw         & 7.45 | 13.37    &   0.36 | 0.24   &      &      \\
\midrule

& & \textbf{WER(\%) zh | en} & \textbf{SIM zh | en} & \textbf{RAE(\%) zh | en} &\\
\cmidrule(lr){3-3} \cmidrule(lr){4-4} \cmidrule(lr){5-5}
\multirow{1}{*}{\textbf{Volume Alteration}}
& \modelraw         & 1.71 | 1.35    & 0.86 | 0.80   &  14.9 | 11.7   &      \\
\midrule

& & \textbf{WER(\%) } & \textbf{SIM} & \textbf{DNSMOS OVRL} &\\
\cmidrule(lr){3-3} \cmidrule(lr){4-4} \cmidrule(lr){5-5}
\multirow{1}{*}{\textbf{Add Sound}}
& GroundTruth & 3.68 & - & 2.12 \\
& \modelraw         &  3.59    & 0.78   &  2.46   &      \\
\midrule

& &\textbf{Model Type} &\textbf{DNSMOS OVRL} & \textbf{DNSMOS SIG} & \textbf{DNSMOS BAK} \\
  \cmidrule(lr){2-6}
\multirow{8}{*}{\textbf{DNS Challenge}}
& FullSubNet \cite{9414177}  &   \multirow{6}{*}{specialized}    &    2.93  &  3.05    &  3.51 \\
& Inter-Subnet \cite{10094858} &       &  2.98  &  3.17    &  3.15 \\
& CDiffuSE \cite{9746901}   &     &  2.84  &  3.37    &  3.52 \\
& SGMSE \cite{10149431}     &     &  3.11  &  3.47    &  3.41 \\
& StoRM \cite{10180108}      &     &  3.15  &  3.54    &  3.69 \\
& GenSE \cite{yao2025gensegenerativespeechenhancement}  &         &  \textbf{3.43}  &  \textbf{3.65}    &  \textbf{4.18} \\
 \cmidrule(lr){2-6}
& MiMo-Audio \cite{coreteam2025mimoaudio}   &   \multirow{2}{*}{general}  &    {3.30}  &  3.56    &  {4.10}    \\
& \modelraw    &     &    3.26  &  {3.59}    &  3.97    \\

\bottomrule
\end{tabular}
} 
\caption{Performance comparison on \editbenchmark{}. The best results are in \textbf{bold}.}
\label{tab:main_results_edit}
\end{table*}

\section{Conclusion}
\label{sec:Conclusion}

Through a multistage training process on the audio tokenizer encompassing reconstruction, feature distillation, and semantic distillation, we have developed a continuous unified representation for speech understanding and generation. In joint training tasks, our model achieves performance on par with leading open-source models, notably reaching state-of-the-art (SOTA) results in the semantic stability of speech generation.

Building upon this unified representation and pre-trained model, we have explored free-form, instruction-based editing that simultaneously handles both semantic and acoustic tasks, showcasing the significant potential of this paradigm. Furthermore, we are open-sourcing our unified tokenizer, pre-trained model, and a speech editing benchmark, inviting more researchers to join us in exploring this promising direction.

\section{Future Work}
\enlargethispage{2\baselineskip}
This report outlines our technical approach and preliminary results. In the following, we summarize several directions that we plan to pursue in the near future.

\begin{itemize}

\item \textbf{Semantic Distillation}: improve large language model distillation to enpower understanding performance.
\item \textbf{Instruction-following generation}: we consider this capability to be a crucial prerequisite for more advanced general-purpose instruction-based speech editing.
\item \textbf{More modalities}: integrated with modalities of images and videos to achieve multimodal understanding and generation.
\end{itemize}


\newpage
\section{Contributors}
\label{sec:contri}

\large{Authors are listed \textbf{alphabetically by the first name}.}

\large{
\begin{multicols}{3}
\raggedcolumns
Ant Inclusion AI \\
Canxiang Yan \\
Chunxiang Jin \\
Dawei Huang \\
Haibing Yu \\
Han Peng \\
Hui Zhan \\
Jie Gao \\
Jing Peng \\
Jingdong Chen \\
Jun Zhou \\
Kaimeng Ren \\
Ming Yang \\
Mingxue Yang \\
Qiang Xu \\
Qin Zhao \\
Ruijie Xiong \\
Shaoxiong Lin \\
Xuezhi Wang \\
Yi Yuan \\
Yifei Wu \\
Yongjie Lyu \\
Zhengyu He  \\
Zhihao Qiu \\
Zhiqiang Fang \\
Ziyuan Huang

\end{multicols}}

\clearpage

\newpage

\bibliographystyle{assets/plainnat}
\bibliography{paper}

\newpage
\beginappendix
\section{Speech Generation}

\begin{figure}
    \centering
    \includegraphics[width=1\linewidth]{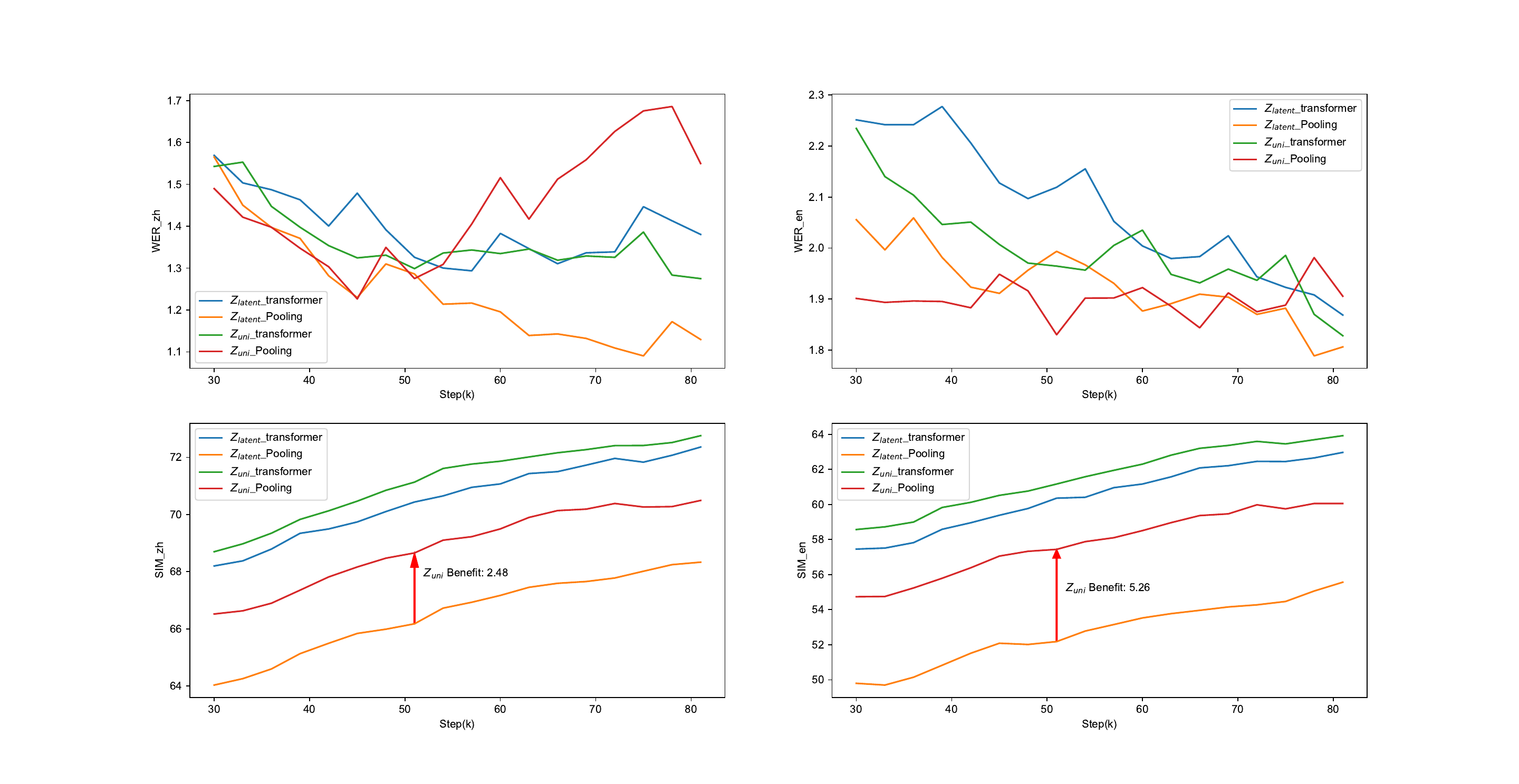}
    \caption{Similarity benefits with $Z_{\text{uni}}$.}
    \label{fig:agg_pool}
\end{figure}

\paragraph{\textbf{Continuous Speech Tokenization}}
\mingaudiotokenizer{} originally extracts continuous speech tokens at 50Hz, which proved redundant for speech generation task. We therefore down‐sampling the tokens to 10Hz. We experimented with Pooling and transformer architecture. Both $Z_{\text{latent}}$ and $Z_{\text{uni}}$ experiments have shown that the transformer architecture outperformed. Moreover, the negative effect of pooling on $Z_{\text{uni}}$ was substantially smaller than that on $Z_{\text{latent}}$, demonstrating the greater robustness of our unified representation. The pooled $Z_{\text{uni}}$ was used for joint training to achieve better understanding performance and faster training speed. As shown in Figure \ref{fig:agg_pool}(smooth=0.6), the similarity with $Z_{\text{uni}}$ is higher than that with $Z_{\text{latent}}$.

\paragraph{\textbf{Per-Token Generation Head}}
To synthesize high-quality audio, we employ a Flow-Matching objective along the optimal-transport (OT) path. Furthermore, we implement classifier-free guidance (CFG) by randomly dropping the LLM conditioning with probability 0.1 during training. Additionally, as shown in Table \ref{tab:dit_mlp}, MLP with higher computational efficiency is also a good choice.

\begin{table}[h]
    \small
    \centering
    \resizebox{0.8\textwidth}{!}{
    \begin{tabular}{lccccc}
        \hline
        \multirow{2}{*}{\textbf{Size}} & \multirow{2}{*}{\textbf{Step$\downarrow$}} & \multicolumn{3}{c}{\textbf{Gen WER(\%) $\downarrow$ | SIM$\uparrow$}} & \multirow{2}{*}{\textbf{RTF(H20)$\downarrow$ }} \\
        \cmidrule(lr){3-5}
        & & \footnotesize \textbf{Seed-zh} & \footnotesize \textbf{Seed-en} & \footnotesize \textbf{AVG} \\
        \hline
        DiT(0.1B) & 378k & 1.040 | \textbf{0.748} & \textbf{1.538} | \textbf{0.682} & \textbf{1.289} | \textbf{0.715} & 0.61\\
        MLP(1.2B) & \textbf{174k} & \textbf{1.032} | 0.733 & 1.910 | 0.654 & 1.471 | 0.694 & \textbf{0.41} \\
        \hline
    \end{tabular}
    }
    \caption{Comparison of the flow matching head size.}
    \label{tab:dit_mlp}
\end{table}

\section{Open-source Audio Data}
\label{sec:app_audio_data}
We include the complete list of open source audio data we used during our training in Table~\ref{tab:audio_data}.

\begin{table*}[bpth]
    \centering
    \small
    \caption{\centering{The complete list of open-source audio data used during our training.}}
    \begin{tabular}{c|c}
    \hline
         Dataset & Audio Length (hrs) \\
         \hline
        WenetSpeech~\citep{zhang2022wenetspeech10000hoursmultidomain}	& 10518	\\
        KeSpeech~\citep{tang2021kespeech}	& 1428	\\
        AliMeeting~\citep{AliMeeting} & 120	\\
        AISHELL-1~\citep{AISHELL1}	& 155	\\
        AISHELL-3~\citep{shi2020aishell}	& 65	\\
        AISHELL-4~\citep{fu2021aishell}	& 61	\\
        CoVoST~\citep{wang2020covost}	    & 456	\\
        CoVoST2~\citep{CoVoST2}	    & 18	\\
        Magicdata~\citep{MagicData_RAMC}	& 747	\\
        Gigaspeech~\citep{GigaSpeech}	& 10288	\\
        Libriheavy~\citep{Libriheavy}	& 51448	\\
        LibriSpeech~\citep{Librispeech}	& 960	\\
        SlideSpeech~\citep{SlideSpeech}	& 473	\\
        SPGISpeech~\citep{SPGISpeech}	& 5000	\\
        TED-LIUM~\citep{rousseau2012ted}	& 208	\\
        Emilla~\citep{he2024emiliaextensivemultilingualdiverse}	    & 90305	\\
        Multilingual LibriSpeech~\citep{pratap2020mls} & 45000 \\
        Peoples Speech~\citep{galvez2021people} &	30000 \\
        \hline
    \end{tabular}
    \label{tab:audio_data}
\end{table*}


\newcommand{\examplebox}[3]{%
  \begin{tcolorbox}[
    breakable,
    enhanced,
    colback=gray!3,
    colframe=gray!60,
    boxrule=0.5pt,
    arc=2mm,               
    left=8pt,right=8pt,top=6pt,bottom=6pt
  ]
    \textbf{Instruction:} #1\par\smallskip
    \textbf{Original text:} #2\par\smallskip
    \textbf{Edited text:} #3
  \end{tcolorbox}\medskip
}

\newenvironment{instrbox}[1]{%
  \begin{tcolorbox}[
    breakable,
    enhanced,
    colback=blue!2,
    colframe=blue!40!black,
    boxrule=0.5pt,
    arc=2mm,
    left=8pt,right=8pt,top=6pt,bottom=6pt,
    title={#1},
    fonttitle=\bfseries
  ]
}{\end{tcolorbox}\medskip}

\section{\textbf{\editbenchmark}\ Examples}
\label{sec:ap_exampels}

\subsection{Examples of Basic Semantic Test Sets}

\subsubsection{Deletion-basic}
\examplebox{delete 'The second for'}{The second for no better reason than the first}{no better reason than the first}

\begin{CJK}{UTF8}{gbsn}
\examplebox{delete the characters or words from index 6 to index 10}{火精灵又出现了猴腿帕克挠挠头}{火精灵又出克挠挠头}

\subsubsection{Insertion-basic}
\examplebox{insert 'publicly' after the character or word at index 3}{It does not comment on specific disputes}{It does not publicly comment on specific disputes}

\examplebox{insert '的行为很可笑' at the end}{在他看来像这种用一排大炮来歼灭一只小虫}{在他看来像这种用一排大炮来歼灭一只小虫的行为很可笑}

\subsubsection{Substitution-basic}
\examplebox{substitute 'a' with 'an outdoor'}{A man wearing a jacket hat and jeans at a market}{A man wearing a jacket hat and jeans at an outdoor market}

\examplebox{substitute the characters or words from index 1 to index 2 with '秋天'}{夏天的酷热过去了冬天的寒冷还早呢}{秋天的酷热过去了冬天的寒冷还早呢}
\end{CJK}

\subsection{Examples of Full Semantic Test Sets}

\subsubsection{Deletion}
\examplebox{Delete the first 4 words.}{Roaming endlessly around the park she wants to go home}{park she wants to go home}

\begin{CJK}{UTF8}{gbsn}
\examplebox{删除“当我们做体验时”}{当我们做体验时我们到底在做什么}{我们到底在做什么}
\end{CJK}

\subsubsection{Insertion}
\examplebox{Start the sentence with `In practice,'}{To turn this into an algorithm only finitely many frequencies are solved for}{In practice, to turn this into an algorithm only finitely many frequencies are solved for}

\begin{CJK}{UTF8}{gbsn}
\examplebox{在句末添加“大家都很感动”}{比如当有人替受伤的小鸟包扎时}{比如当有人替受伤的小鸟包扎时大家都很感动}
\end{CJK}

\subsubsection{Substitution}
\examplebox{Change the last two words to `accidentally'}{The boy wanted to believe that his friend had simply become separated from him by accident}{The boy wanted to believe that his friend had simply become separated from him accidentally}

\begin{CJK}{UTF8}{gbsn}
\examplebox{把“不变应万变”换成“静观其变”}{采取以不变应万变的不卖不赔方法}{采取以静观其变的不卖不赔方法}
\end{CJK}

\subsection{Instruction Examples of Acoustic Test Sets}

\begin{instrbox}{Dialect conversion Instructions}
\begin{itemize}[leftmargin=*, itemsep=2pt]
  \item Change the accent of the speech to Chengdu.
  \item Change the accent of the speech to Dongbei.
  \item Change the accent of the speech to Guangxi.
\end{itemize}
\end{instrbox}

\begin{instrbox}{Speech changing Instructions}
\begin{itemize}[leftmargin=*, itemsep=2pt]
  \item Adjusts the speed to 0.5
  \item Adjusts the speed to 1.7
  \item Adjusts the speed to 1.3
\end{itemize}
\end{instrbox}

\begin{instrbox}{Pitch changing Instructions}
\begin{itemize}[leftmargin=*, itemsep=2pt]
  \item Shifts the pitch by -4 steps
  \item Shifts the pitch by 6 steps
  \item Shifts the pitch by -3 steps
\end{itemize}
\end{instrbox}

\begin{instrbox}{Volume changing}
\begin{itemize}[leftmargin=*, itemsep=2pt]
  \item Adjusts the volume to 0.5
  \item Adjusts the volume to 0.3
  \item Adjusts the volume to 1.5
\end{itemize}
\end{instrbox}

\end{document}